\documentclass[a4paper,conference]{IEEEtran}
\IEEEoverridecommandlockouts
\usepackage{stfloats,color}
\usepackage{amsfonts}
\usepackage{amsmath}
\usepackage{dsfont}
\usepackage{setspace}
\usepackage{adjustbox}
\usepackage{ragged2e}
\usepackage{cite}
\usepackage{bbm}

\usepackage{filecontents}
\usepackage{array}

\usepackage{mdwmath}
\usepackage{multirow}
\usepackage{makecell}
\usepackage{times}
\usepackage{diagbox}
\usepackage{epsfig}
\usepackage{latexsym}
\usepackage{epstopdf}
\usepackage{verbatim}
\usepackage{units}
\usepackage{amsthm}
\usepackage{mdwlist}
\usepackage{esvect}
\usepackage{tabularx,colortbl}
\usepackage[unicode=true, linktocpage, linkbordercolor={0.5 0.5 1}, citebordercolor={0.5 1 0.5}, linkcolor=blue]{hyperref}

\begin{document}

% \title{Turning Brain MRI into Diagnostic PET: $^{15}$O-water PET Synthesis from Multi-contrast MRI via Attention-based Encoder–Decoder Networks}
% \title{Attention-guided encoder decoder networks for Brain PET Synthesis from Multi-contrast MRI for cerebrovascular disease assessment}
% \title{Turning MRI into PET: $^{15}$O-water Brain PET Synthesis from Multi-contrast MRI via Attention-based Encoder–Decoder Networks}
% \title{Optimized encoder decoder network with attention mechanisms to synthesize PET from Multi-contrast MRI} 
% \title{MRI-to-PET translation via attention-based encoder-decoder networks}
% \title{Brain PET Synthesis from Multi-contrast MRI via Attention-Based Encoder-Decoder Networks}
% \title{Brain MRI-to-PET Synthesis with Multi-Attention Convolutional Neural Network}
\title{Brain MRI-to-PET Synthesis using 3D~Convolutional Attention Networks}

\author{\IEEEauthorblockN{
		Ramy Hussein\IEEEauthorrefmark{1}\IEEEauthorrefmark{5},
		David Shin\IEEEauthorrefmark{2},
		Moss Zhao\IEEEauthorrefmark{1},
		Jia Guo\IEEEauthorrefmark{3},
		Guido Davidzon\IEEEauthorrefmark{4},
		Michael Moseley\IEEEauthorrefmark{1},
		Greg Zaharchuk\IEEEauthorrefmark{1} 		 
	}
	\IEEEauthorblockA{\IEEEauthorrefmark{1}Radiological Sciences Laboratory, Stanford University, Stanford, CA  94305, USA}
	\IEEEauthorblockA{\IEEEauthorrefmark{2}Global MR Applications \& Workflow, GE Healthcare, Menlo Park, CA 94025 , USA}
	\IEEEauthorblockA{\IEEEauthorrefmark{3}Department of Bioengineering, University of California Riverside, CA 92521, USA}	
	% \IEEEauthorblockA{\IEEEauthorrefmark{4}Division of Nuclear Medicine, Department of Radiology, Stanford University, Stanford, California 94305, USA}
	\IEEEauthorblockA{\IEEEauthorrefmark{4}Nuclear Medicine \& Molecular Imaging, Stanford University, Stanford, California 94305, USA}
	\IEEEauthorblockA{\IEEEauthorrefmark{5}Corresponding author: ramyh@stanford.edu}
}

\maketitle

\begin{abstract}

Accurate quantification of cerebral blood flow (CBF) is essential for the diagnosis and assessment of a wide range of neurological diseases. Positron emission tomography (PET) with radiolabeled water ($^{15}$O-water) is considered the gold-standard for the measurement of CBF in humans. PET imaging, however, is not widely available because of its prohibitive costs and use of short-lived radiopharmaceutical tracers that typically require onsite cyclotron production. Magnetic resonance imaging (MRI), in contrast, is more readily accessible and does not involve ionizing radiation. This study presents a convolutional encoder-decoder network with attention mechanisms to predict gold-standard $^{15}$O-water PET CBF from multi-sequence MRI scans, thereby eliminating the need for radioactive tracers. Inputs to the prediction model include several commonly used MRI sequences (T1-weighted and T2-FLAIR, and arterial spin labeling). The model was trained and validated using 5-fold cross-validation in a group of 126 subjects consisting of healthy controls and cerebrovascular disease patients, all of whom underwent simultaneous $^{15}$O-water PET/MRI. The results show that such a model can successfully synthesize high-quality PET CBF measurements (with an average SSIM of 0.924 and PSNR of 38.8 dB) and is more accurate compared to concurrent and previous PET synthesis methods. We also demonstrate the clinical significance of the proposed algorithm by evaluating the agreement for identifying vascular territories with abnormally low CBF. Such methods may enable more widespread and accurate CBF evaluation in larger cohorts who cannot undergo PET imaging due to radiation concerns, lack of access, or logistic challenges.
\end{abstract}

\IEEEpeerreviewmaketitle

%%%%%%%%%%%%%%%%%%%%%%%%%%%%%%%%%
\section{Introduction}
\label{sec1}

Cerebrovascular diseases are a major cause of death globally, affecting all racial and ethnic groups \cite{yusuf2001global}. Cerebrovascular diseases include stroke, aneurysm, and Moyamoya syndrome. Stroke, for example, affects around 15 million people worldwide annualy \cite{mukherjee2011epidemiology}. Of these, five million die, and another five million are left permanently disabled, placing a burden on family, community, and healthcare systems. The prompt detection and adequate evaluation of cerebrovascular diseases leads to faster treatment and less damage to the brain. Furthermore, many other neurological diseases are characterized by or associated with abnormalities in cerebral blood flow (CBF), including vascular malformations, seizure disorders, and neurodegenerative conditions such as Alzheimer’s disease. For these reasons, accurate CBF quantification is crucial for the diagnosis and assessment of cerebrovascular disorders. 

Positron emission tomography (PET) with radiolabeled water ($^{15}$O-water) is widely considered the gold-standard imaging technique for measuring CBF in humans. PET, however, is not widely available because of its prohibitive costs, difficult logistics, and use of ionizing radiation. In fact, probably less than 20  centers in the world can perform $^{15}$O-water PET CBF imaging, and most of these sites can perform studies only in the research setting. Magnetic resonance imaging (MRI), on the other hand, is less expensive, far more readily available, and does not require radioactive tracers. The two most common MRI exams used to quantify CBF are arterial spin labeling (ASL) perfusion MRI and dynamic susceptibility contrast (DSC) perfusion MRI \cite{muir2014quantitative}. Both types of MRI exams are widely used clinically and in neuroscience research applications in recent years \cite{wu2010quantification}. Nevertheless, the quantification of MRI-derived CBF maps can be inaccurate in the presence of global or focal CBF reductions, as occurs frequently in patients with suspected cerebrovascular diseases (\textit{e.g.}, carotid artery disease or decreased cardiac output) \cite{grade2015neuroradiologist}. This has motivated researchers to develop image-to-image translation methods to synthesize PET-like CBF maps from MRI scans. Such methods can potentially improve the quantitative and qualitative assessment of CBF when compared to perfusion MRI-derived CBF measurements, and would find use in a wider range of patients and indications than is feasible with PET imaging.

Aided by the increasing size of medical imaging databases and recent advances in the computer vision field, image-to-image translation methods using deep learning have been developed that can transform one medical image modality to another. This capability has been shown for predicting computed tomography (CT) images from MRI \cite{kearney2020attention}, MRI from CT \cite{jin2019deep}, CT from PET \cite{armanious2019unsupervised}, PET from CT \cite{ben2019cross}, MRI from PET \cite{bazangani2022fdg}, as well as different MRI contrasts from one another \cite{2019dar}. It should be noted that the clinical utility of such mappings has not been established. In the last few years, several image synthesis methods were introduced to transform multi-parametric brain MRI images into PET CBF maps \cite{2020guo, 2021yousefi}. These methods can potentially extend the ability to quantitatively characterize cerebrovascular disorders to sites without the capability to perform the gold-standard PET imaging. For instance, \textit{Guo et al.} \cite{2020guo} used a deep convolutional neural network (CNN) to predict $^{15}$O-water PET scans from multi-contrast MRI inputs, achieving an average structural similarity index (SSIM) of 0.85 in both normal subjects and patients with cerebrovascular disease. In another study \cite{2021yousefi}, an attention-guided CNN was introduced for translating T1-weighted and ASL data to PET-like images; achieving a comparable SSIM of 0.85 in normal subjects. 

Although existing MRI-to-PET translation models could synthesize PET images, they either produced imprecise CBF measurements or failed to attain clinical applicability due to lack of generalization capacity. These earlier works employed relatively basic architectures (simple U-Net and Pix2Pix) and we believe more advanced methodologies could lead to better performance, in particular, to make them more clinically generalizable to a wider range of disease conditions. In this work, we introduce a multimodal encoder-decoder attention network to synthesize high-quality PET CBF maps from structural and perfusion MRI scans. The inputs to the network include structural (T1w and T2-FLAIR) and perfusion images (single-delay [SD] and multi-delay [MD] ASL), as well as CBF estimates derived from the ASL sequences. We performed experiments to evaluate the model performance with different loss functions, network settings, and subsets of input MRI scans. The experimental results demonstrate the effectiveness of the proposed method, achieving superior PET prediction performance when compared to the state-of-the-art methods. The main contributions of this work can be summarized as follows:

\begin{itemize}
	\item A 3D convolutional encoder-decoder network based on attention mechanisms is proposed. An efficient loss function is developed and described.
	
	\item In comparison to previous MRI-to-PET translation methods, our model yields superior performance with an average SSIM of 0.924 and peak signal-to-noise ratio (PSNR) of 38.80dB. Best performance was seen with the presence of attention mechanisms, using a custom loss function defined as a weighted sum of mean absolute error (MAE) and SSIM. 
	
	\item The quantitative results demonstrate that integrating anatomical and tissue perfusion information from structural and ASL MRI exams can notably improve the prediction of PET CBF for both healthy controls and patients with cerebrovascular diseases. 
	
	\item We describe and evaluate using receiver operator characteristic methodology a clinically meaningful metric to assess the quality of the predicted images: the ability to identify brain regions with abnormally low CBF.

	\item Qualitative and paired comparison analyses show strong correlation and high levels of agreement between regional CBF values in synthetic and ground truth PET scans. The vascular territories with impaired CBF can be accurately identified in synthetic PET scans, leading to better diagnosis and assessments of cerebrovascular diseases at MRI-only sites.
	
	% \item We have derived PET-like CBF maps from patients who underwent MRI only as part of their clinical care, demonstrating the utility of this method to produce gold-standard CBF outside of traditional PET settings.

\end{itemize}

\section{Related Work}

In this section, we first review the concurrent and previous deep learning models that are commonly used in cross-modality brain image synthesis applications. We then review the recent MRI-to-PET translation networks that are related to our study.

\subsection{Cross-modality brain image-to-image translation}

Recently, image-to-image translation networks have been proposed to solve various image prediction problems in the medical field. For instance, Wolterink \textit{et al.} \cite{2017wolterink} took a pioneering role in applying deep learning for cross-modality medical image synthesis. They used a generative adversarial network (GAN) to convert 2D brain MRI images into 2D brain CT images and vice versa. Quantitative results using a separate test set of six patients showed that GANs can reasonably produce CT images that closely mimic the appearance of actual CT images, achieving an average PSNR of 32.3dB. In \cite{2019dar}, Dar \textit{et al.} employed conditional generative adversarial networks for multi-contrast MRI synthesis. They used the adversarial loss together with the pixel-wise and perceptual losses which helped improve the synthesis performance of both T1- and T2-weighted images. In \cite{2020yangQ}, Yang \textit{et al.} developed cross-modality MRI image generation method for multimodal registration and segmentation using conditional generative adversarial networks. This method achieved comparable results on five brain MRI datasets while using a single modality image as an input. 

In \cite{2020armanious}, a GAN-based framework with a new generator architecture and style-transfer losses was introduced to address three different medical image-to-image translation problems: PET-to-CT translation, MRI motion artefacts correction, and PET image denoising. The quantitative results and radiologists' evaluations for the three tasks demonstrate the superiority of the proposed GAN architecture over the existing translation methods. In \cite{2020yang}, Yang \textit{et al.} introduced a structure-constrained cycleGAN for unsupervised MRI-to-CT synthesis. By using an additional structure-consistency loss function together with a self-attention module, the cycleGAN approach was able to produce high-quality synthetic brain and abdomen CT images. In \cite{2022liu}, Liu \textit{et al.} proposed to use a transformer-based MRI synthesis approach, named multi-contrast multi-scale transformer (MMT), for missing MRI sequence imputation. Their proposed MMT network was able to efficiently take any subset of input MRI contrasts and synthesize those were missing. Experiments on two multi-contrast MRI datasets showed that MMT can quantitatively and qualitatively outperform the state-of-the-art MRI synthesis methods. The qualitative evaluations also revealed that vision transformers can be used, not only for medical image recognition problems, but also for the more challenging image-to-image translation problems.

Image-to-image translation networks were also used to reduce or even eliminate the need for gadolinium-based contrast agents (GBCAs) in MRI studies. In \cite{2018gong}, Gong \textit{et al.} employed a convolutional encoder-decoder network to synthesize diagnostic quality contrast-enhanced MRI~(CE-MRI) images from images with only 10\% of the full contrast agent dose level, Quantitative results and qualitative assessments showed that the synthesized post-contrast MRI images have significant improvements in image quality and contrast enhancement when compared to the acquired low-dose images. In \cite{2021chen}, a 3D high-resolution fully convolutional network was designed to map a set of pre-contrast MRI scans to CE-MRI. The pre-contrast MRI sequences of T1-weighted (T1w), T2-weighted (T2w), and apparent diffusion coefficient (ADC) map were used as inputs to the network and the post-contrast T1w being the target output. Results showed high-quality synthetic contrast-enhanced MRI images, potentially allowing deep learning to substitute for gadolinium contrast agents and decrease gadolinium deposition. 

Others \cite{2021preetha} have tried to address the same synthesis problem by using T1w, T2w, and fluid-attenuated inversion recovery (FLAIR) MRI sequences as inputs to two deep convolutional neural networks (dCNN). Results showed that the quantification of the synthetic CE-MRI images can efficiently allow the assessment of the patient’s response to treatment with insignificant difference when compared to true CE-MRI images collected with gadolinium administration. Furthermore, Xie \textit{et al.} \cite{2022xie} developed an advanced cascade neural network architecture that incorporates the contour information of the input unenhanced MRI images to improve the quality of the synthetic CE-MRI images. Quantitative and qualitative assessments on a hold-out test set of 169 patients showed that, thanks to the contour information, there was no intensity differences observed in both tumor and non-tumor brain regions. 

\subsection{MRI-to-PET translation}

The feasibility of transforming MRI data into PET-like images were explored by several research groups. For instance, Sikka \textit{et al.} \cite{2018sikka} used 3D U-Net architecture to transform MRI scans to FDG-PET scans and then integrated both to improve the diagnosis of Alzheimer’s disease. The use of the synthesized PET data yielded an average increase in the classification accuracy of 4.25\%. In \cite{2020lan}, Lan \textit{et al.} developed a 3D self-attention conditional GAN (SC-GAN) that adopted attention modules to forge the relationships between different neuroimaging modalities. The performance of the SC-GAN was examined on the ADNI database, where the SC-GAN was used to take multimodal MRI as input to synthesize several different downstream image contrasts, including amyloid PET as well as fractional anisotropy (FA) and mean diffusivity (MD) maps. Although the quantitative results showed high synthesis accuracy, the synthesis error was found to be relatively high in the brain regions with high amyloid-$\beta$ load, and the authors concluded that the synthetic PET could not replace amyloid PET imaging for clinical tasks. 

Chen \textit{at al.} \cite{2019chen} employed a 2D U-Net model to predict full-dose amyloid PET images from a combination of synthesized or acquired extremely low-dose amyloid PET and multi-contrast MR images. Quantitative analyses showed that this type of imaging integration can effectively produce synthetic PET images with standardized uptake value ratio (SUVR) values that are comparable to those of the actual full-dose PET images. Guo \textit{et al.} \cite{2020guo} showed the feasibility of synthesizing $^{15}$O-water PET CBF images solely from multi-contrast MRI scans. They adopted a U-Net-structured CNN that takes structural MRI and single- and multi-delay ASL exams as inputs to predict a simultaneously acquired PET CBF map, allowing more accurate CBF measurements in MRI-only sites. Yousefi \textit{et al.} \cite{2021yousefi} also developed a CNN-based synthesis approach to convert the ASL to PET CBF images. In addition to ASL, they also incorporated T1w MRI to provide anatomical information to improve PET synthesis. The quantitative measures and a blind reader study showed high levels of similarity between the true and synthetic PET images, achieving an average SSIM of 0.85, though it should be noted that all subject were healthy and no pathology was present in the images. In \cite{2020chen}, Chen \textit{et al.} used a similar deep learning architecture to predict cerebrovascular reserve (CVR, defined as the percent CBF increase from a baseline value after acetazolamide administration) images from structural and perfusion-based MRI scans as well as the baseline $^{15}$O-water PET CBF measured before acetazolamide administration. Quantitative and comparative analyses showed high diagnostic performance in identifying vascular territories with impaired CVR. 

Hu \textit{et al.} \cite{2020hu} used a bidirectional GAN to transform MRI into PET images while preserving the diverse brain structures of different subjects. They demonstrated satisfactory levels of quantitative results but with limited quality of synthetic PET images. In \cite{2020shin}, Shin \textit{et al.} adopted the bidirectional encoder representations from transformers (BERT) algorithm to generate synthetic amyloid and FDG PET images from T1w MRI data with minimal pre- and post-processing. However, they showed limited quantitative and qualitative PET prediction results, making it impracticable in clinical settings. In \cite{wang20183d}, a cGANs-based method was introduced to estimate the full-dose from low-dose PET images alone. Using brain data for healthy subjects and mild cognitive impairment patients, their model showed better quantitative and qualitative performance than baseline methods. 

%%%%%%%%%%%%%%%%%%%%%%%%%%%%%%%%%

\section{Materials and Data Preprocessing}
\label{sec2}

This study was approved by the Institutional Review Board of Stanford University in accordance with the ethical standards of the Helsinki declaration for medical research involving human subjects, and is HIPAA compliant. Written informed consent was obtained from all participants prior to the study. Our dataset was acquired between April 2017 and March 2022. Data were acquired from 131 subjects (72 healthy controls [HC] and 59 cerebrovascular disease patients [PT]) on a 3T PET/MR hybrid system (SIGNA PET/MR, GE Healthcare, Waukesha, WI, USA) using an 8-channel head coil. Patients were instructed to refrain from food or beverage containing caffeine at least 6 hours before imaging.

\begin{figure*}[!ht]
	\centering
	\includegraphics[width=0.95\textwidth]{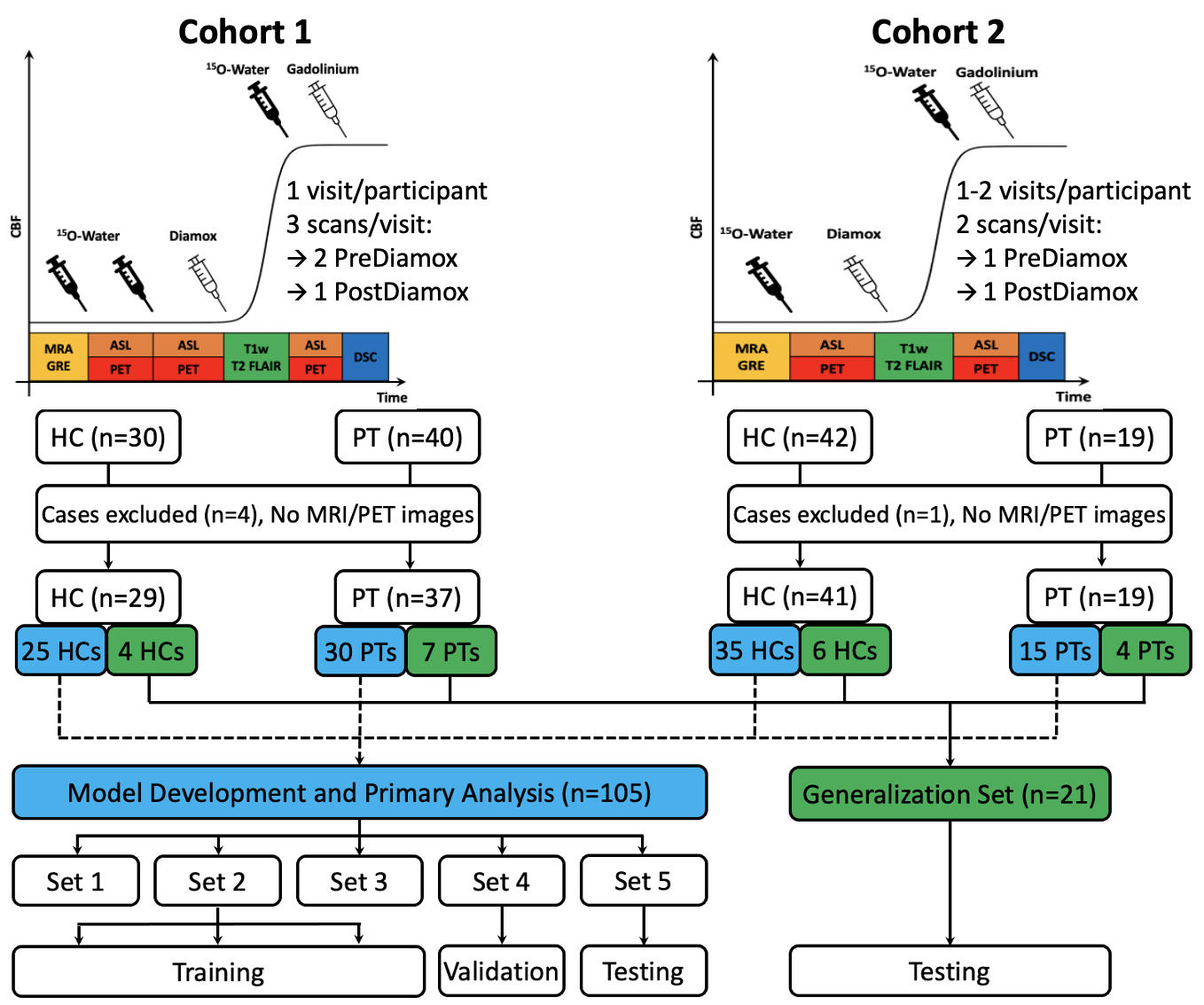}
	\caption{Experimental design for measuring CBF using PET/MRI in two cohorts. In cohort 1, three simultaneous PET/ASL acquisitions were acquired from the participants in a single visit (2 scans before and one scan 15 minutes after the administration of the vasodilator [acetazolamide, ACZ]. In cohort 2, two simultaneous PET/ASL acquisitions were acquired from the participants in each visit (one before and one 15 minutes after ACZ administration); Of the 41 HCs in cohort 2, 31 had two separate imaging sessions on different days.}
	\label{fig.setup}
\end{figure*}

As shown in Figure~\ref{fig.setup}, our dataset consisted of two cohorts. The first cohort included PET/MRI data collected from 30 HCs and 40 PTs, of which 4 cases were excluded due to missing MRI or PET scans. Each participant had only one visit, in which three simultaneous PET/ASL acquisitions were acquired (two before and one 15 minutes after the intravenous administration of the vasodilator [acetazolamide, ACZ] at a dose of 15 mg/kg up to a maximum of 1 g). The second cohort included PET/MRI data acquired from 42 HCs and 19 PTs, of which 1 HC participant was excluded due to missing PET scans. In this cohort, around 75\% of HC participants (31/41) underwent two identical sessions on different days. During each session, two simultaneous PET/ASL acquisitions were acquired from the participants (one before and one 15 minutes after ACZ administration). 

MRI perfusion scans included two ASL acquisitions: single-delay (SD) and multi-delay (MD) pseudo-continuous ASL (pCASL). Dynamic susceptibility contrast (DSC) perfusion MRI was also performed after ACZ administration. MRI-based CBF maps were made from both single-delay and multi-delay ASL using a general kinetic model \cite{buxton1998general, alsop2015recommended}. Arterial transit time (ATT) was derived from the multi-delay ASL using a non-iterative method \cite{dai2012reduced}. Proton density (PD) images were also collected as part of the ASL acquisitions for quantitation purposes. Magnetic resonance angiography (MRA), gradient-echo (GRE), T1-weighted, and T2-weighted fluid-attenuated inversion recovery (T2- FLAIR) were also acquired from all participants. A list of the parameters used for the ASL and structural MR imaging can be found in Table~\ref{tab_MRIparameters}.

In both cohorts, quantitative PET CBF was determined using $^{15}$O-water injection and an image-derived arterial input function (AIF) kinetic model \cite{khalighi2018image} and the 1-compartment model \cite{zhou2001linear} using PMOD software. The PET images were reconstructed with a resolution of 1.56$\times$1.56$\times$2.78 mm$^3$ and were corrected for signal decay and attenuation. All images were co-registered to the T1w images using Statistical Parametric Mapping (SPM) software, and then normalized to the Montreal Neurological Institute (MNI) brain template \cite{mazziotta2001probabilistic} with 2mm isotropic resolution using Advanced Normalization Tools (ANTs) software \cite{tustison2014large}. The brain tissue segmentation was performed using FSL \cite{smith2004advances} and all 3D PET/MRI images were cropped to 96$\times$96$\times$64 voxels for faster computations. 

%%%%%%%%%%%%%%%%%%%%%%%%%%%%%%%%%
\section{Methodology}
\label{sec3}

\subsection{Proposed Network Architecture}

\begin{figure*}[!ht]
	\centering
	\includegraphics[width=\textwidth]{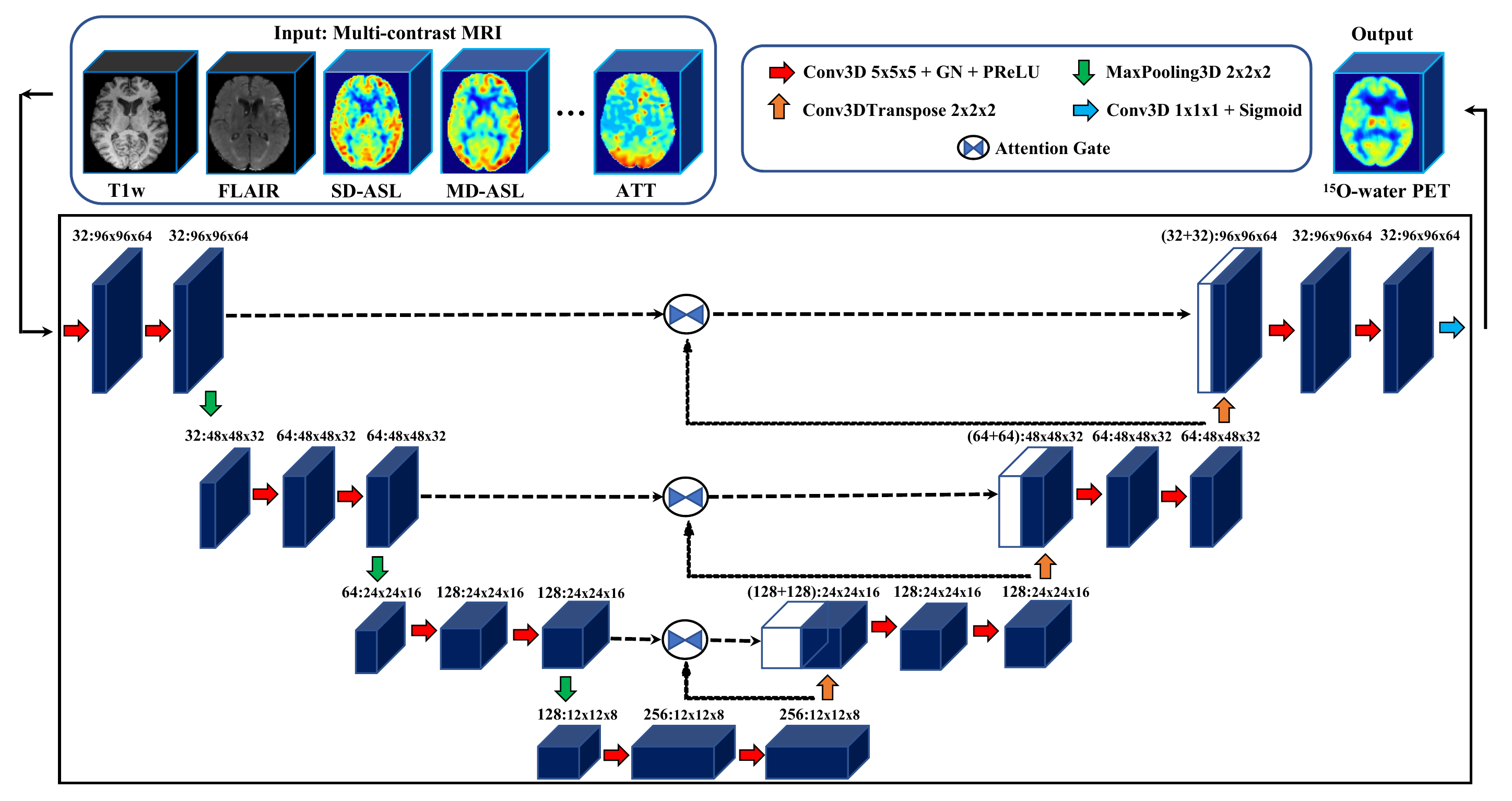}
	\caption{Attention-based encoder-decoder network architecture for predicting PET CBF maps from multi-contrast MRI.}
	\label{fig.model}
\end{figure*}

Figure~\ref{fig.model} shows the architecture of the proposed 3D convolutional encoder-decoder network. The input to the network is an 8-channel tensor $ \mathbf{X} $ $\in$~$\mathbb{R}^{h\times w\times d\times 8}$ that includes data both from structural MRI (T1w, T2-FLAIR) and perfusion MRI scans (perfusion weighted maps from SD-ASL and MD-ASL) as well as the following quantified metrics: SD-CBF, MD-CBF, and ATT derived from MD-ASL. The output of the network $ \mathbf{Y} $~$\in$~$\mathbb{R}^{h\times w\times d\times 1}$ denotes the PET CBF map. To transform multi-contrast MRI into PET, the proposed encoder-decoder network is developed to serve as a non-linear mapping function $f_{\theta}$, such that $ \mathbf{Y} = f_{\theta} \left(  \mathbf{X} \right)$, where $\theta$ contains the network parameters to be learned. 

In a previous study, we showed the ability to predict the gold-standard $^{15}$O-water PET CBF from a set of 16 input MRI contrasts using a 2D convolutional neural network \cite{2020guo}. In this study, we significantly improved the quality of synthetic PET by using an attention-based 3D structure that leveraged the spatial information across eight volumetric MRI scans and captured the long-range feature interactions needed for reliable predictions. The major limitations of applying 3D models to brain image-to-image translation problems are the limited availability of annotated brain imaging data and the associated high computational cost. Therefore, we used a significantly larger cohort of healthy controls and cerebrovascular disease patients for the current study and also applied several data augmentation strategies to further enlarge the overall number of PET/MRI data samples needed for enhanced translation performance. Finally, a custom loss function was carefully designed to preserve contextual and structural information in input multi-contrast MRI scans and thus optimize the performance of the MRI-to-PET translation network.  

As illustrated in Figure~\ref{fig.model}, the proposed attention-guided encoder-decoder network leverages the power of 3D convolutional neural networks, trained end-to-end, to efficiently integrate multiple MRI scans to generate high-quality synthetic PET scans. The network consists of three prime modules: the encoder, decoder, and attention mechanisms. The encoder performs a series of consecutive 3D convolutions to compress the input multimodal MRI volumes into low-resolution representation maps. The decoder, on the other hand, applies 3D deconvolutions to these representation maps and then performs upsampling operations to output the PET CBF maps. Attention mechanisms are also used to search the relevant aspects of the input MRI scans at both channel and spatial levels, leading to a fine-grained quality prediction. To identify the input MRI sequences that have the greatest impact on the PET CBF prediction, the model was trained and tested using structural MRI only, perfusion MRI only (\textit{i.e.}, the ASL sequences), and a combination of structural and perfusion MRI images. 

The proposed convolutional encoder-decoder network was implemented in Python using the TensorFlow framework. The number of filters used in the four convolutional layers of the network encoder is 64, 128, 256, and 512, respectively. The kernel size used for each of the convolutional layers was 5$\times$5$\times$5. Training and testing of the network were carried out on two NVIDIA Tesla V100-PCIE Volta graphics processing units (GPUs). The custom loss function described in Section~\ref{subsec4_4} was used to optimize the weights of the network and improve its predictive power. 

\subsection{Attention Mechanisms}

Convolutional encoder-decoder networks are widely used in image segmentation and image-to-image translation algorithms. However, the use of predefined convolutional filters constrains the encoder-decoder networks from learning the global information while leveraging local information only \cite{zhang2019self}. This results in a non-trivial bias by discarding some of the critical features needed for accurate performance. Using larger convolutional filters and deeper encoder-decoder architectures are examples of the naive solutions introduced to enhance the image segmentation or synthesis performance. These solutions, however, may seriously impact the computational complexity without recognizably improving the results. 

\begin{figure*}[!ht]
	\centering
	\includegraphics[width=\textwidth]{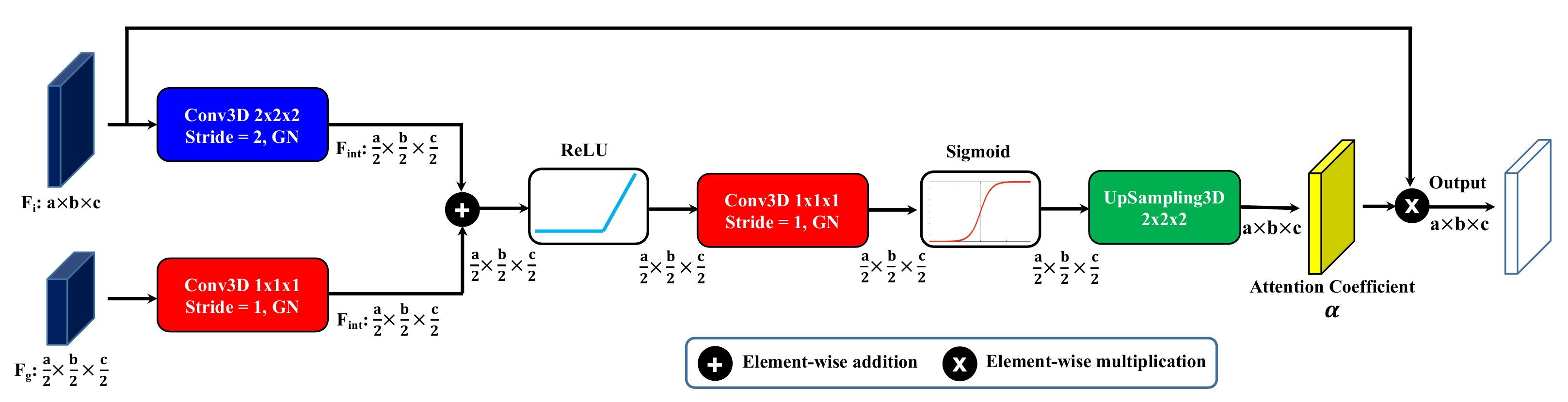}
	\caption{The schematic of an attention mechanism used in the 3D convolutional encoder-decoder network. Input features ($\mathbf{F_i}$) are multiplied element-wise by attention coefficients ($\mathbf{\alpha}$) computed in the attention module. The gating features ($\mathbf{F_g}$) collected from a lower layer of the network are used to identify the spatial regions of interest with relevant activations and contextual information, GN denotes group normalization.}
	\label{fig.attention}
\end{figure*}

Attention mechanisms have been proposed as an advanced solution that captures long-range feature interactions and thus boost convolutional encoder-decoder network performance. In our MRI-to-PET synthesis problem, 3D MRI and PET images include both brain and non-brain voxels (\textit{i.e.}, zero voxels outside the brain margins). Also, for a patient diagnosed with a neurological disorder, abnormal lesions could only be seen in specific regions of interest (ROIs) in certain MRI scans. Therefore, we propose to use the attention mechanism, shown in Figure~\ref{fig.attention} \cite{oktay2018attention}, at the skip connection to enable the encoder-decoder network to focus not only on the brain voxels but also on the ROIs with brain abnormalities of varying size and appearance. 

We chose to use an additive soft attention that works by assigning different weights to different parts of the feature map. Larger weights are assigned to the regions of high relevance and smaller weights go to regions of lower relevance. During the model training, the weights of soft attention are optimized such that the model can better identify which regions to pay more attention to. The attention mechanism takes in two inputs, the input features ($\mathrm{F_i}$) coming from the encoder and the gating features ($\mathrm{F_g}$) coming from a coarser scale of the decoder. Since maximum pooling operations are recursively applied in the downsampling path of the network, $\mathrm{F_g}$ often has smaller dimensions and better feature representation than $\mathrm{F_i}$, given that $\mathrm{F_g}$ comes from deeper layers in the network. The dimensions of $\mathrm{F_i}$ are reduced through a strided 3D convolution producing an intermediate feature map $\mathrm{F_{int}}$. A similar intermediate feature map is produced by applying a regular 1$\times$1$\times$1 3D convolution to the gating features. The two intermediate feature maps are then summed element-wise, where aligned weights get larger and unaligned weights to become relatively smaller. The resultant tensor is fed into a rectified linear unit (ReLU) activation function and then to a 1$\times$1$\times$1 3D convolution and a sigmoid function, producing the attention coefficients ($\alpha$). Lastly, $\alpha$ is upsampled to the original dimensions of $\mathrm{F_i}$ and then multiplied element-wise to $\mathrm{F_i}$, scaling the different regions of the input feature map according to their relevance. 

Since different MRI sequences and spatial patterns impose different effects on the quality of synthetized PET images, an additive soft attention mechanism was embedded into the encoder-decoder network to concurrently search the relevant aspects of the input at the channel and spatial levels for a fine-grained quality prediction. 

\subsection{Image Quality Assessment}

The similarity between the synthetic and reference PET CBF images were evaluated using root-mean-square error (RMSE), structural similarity index (SSIM), and peak signal-to-noise ratio (PSNR), defined as:

%\small
%\begin{equation}
%	\small
%	NRMSE  = \dfrac{1}{x_{max}-x_{min}} \sqrt {\dfrac{1}{mnp} \sum\limits_{i=1}^{m} \sum\limits_{j=1}^{n} \sum\limits_{k=1}^{p}  \Big( x(i,j,k)- y( i,j,k)   \Big) ^2 }
%\end{equation}	
%\normalsize

\begin{equation}
	% \small
	RMSE  = \sqrt {\dfrac{1}{mnp} \sum\limits_{i=1}^{m} \sum\limits_{j=1}^{n} \sum\limits_{k=1}^{p}  \Big( x(i,j,k)- y( i,j,k)   \Big) ^2 }
\end{equation}	

where $x$ and $y$ refer to the reference and synthetic PET scans, $x_{max}$ and $x_{min}$ are the maximum and minimum intensity values of the reference PET, $x(i,j,k)$ and $y(i,j,k)$ are the reference and predicted voxel intensity values, and $m$, $n$, and $p$ are the dimensions of reference/predicted PET images.	
	
\begin{equation}
	% \small
	SSIM  =  \dfrac{  \left( 2 \mu_x \mu_y + c_1 \right) \left(  2 \sigma_{xy} + c_2 \right) }{    \left( \mu_x^2 + \mu_y^2  + c_1\right) \left(  \sigma_x^2 + \sigma_y^2  + c_2 \right)  } 
\end{equation}
where $\mu_x$ and $\mu_y$ denote the mean values of reference and synthetic PET images, $\sigma_x^2$ and $\sigma_y^2$ denote the variance of reference and synthetic PET images, $\sigma_{xy}$ is the covariance of both PET images, $c_1$ and $c_2$ are two constants used to stabilize the division with weak denominator, $c_1=(k_1 L)^2$, $c_2=(k_2 L)^2$, $k_1=0.01$, $k_2=0.03$, and $L$ is the dynamic range for the voxel intensity values of PET images. 

\begin{equation}
	% \small
	PSNR = 10 \cdot \log_{10} \Bigg( \dfrac{y_{max} \cdot  mnp}{ \sum\limits_{i=1}^{m} \sum\limits_{j=1}^{n} \sum\limits_{k=1}^{p}  \Big( x(i,j,k)- y( i,j,k)   \Big) ^2} \Bigg)
\end{equation}
where $y_{max}$ is the maximum voxel intensity value of the image.

\subsection{Custom Loss Function}
\label{subsec4_4}

The loss function in medical image translation problems is a measure of how accurately the predictive model can replicate the target image. In our MRI-to-PET translation problem, the loss function took two inputs: the reference PET image and the synthetic PET image produced by our encoder-decoder network. Lower values of the loss indicate more accurate prediction performance and higher values indicate poor performance. The mean squared error (MSE) is one of the most commonly used loss functions in image synthesis problems. MSE, however, was found to be ineffective in processing medical images that are prone to artifacts and ghosting. In addition, MSE treats the different regions of a medical image equally, which might result in fairly good overall predictions but a few very poor predictions in the regions of most interest (\textit{e.g.}, abnormal lesions that are crucial for the disease assessment). This motivated the development of more appropriate loss functions that better characterize the structure of the synthesized PET with respect to the reference PET scan.

In this study, a carefully designed loss function is introduced to optimize the prediction performance of the proposed encoder-decoder network. The proposed loss function is a combination of multiple loss components that work cooperatively to drive the network toward the most representative distribution of actual PET images.

\textit{1) Voxel-wise reconstruction loss:} The mean absolute error (MAE) is used as a reconstruction loss that measures the voxel-wise similarity between real and predicted images, and it is defined as:
	
\begin{equation}
	% \small
	\mathcal{L}_{r}  =  \dfrac{1}{mnp} \sum\limits_{i=1}^{m} \sum\limits_{j=1}^{n} \sum\limits_{k=1}^{p}  \Big| x(i,j,k) - y(i,j,k)   \Big|
\end{equation}
where $x(i,j,k)$ and $y(i,j,k)$ are the voxel intensity values of reference and synthetic PET images. 

The use of the reconstruction loss is twofold. First, it is adopted to minimize the difference between target and predicted images at a voxel level. It also serves as a regularizer that helps the network maintain robust prediction performance when applied to data with outliers and artifacts.  

\textit{2) Perceptual loss:} In addition to the reconstruction loss, the SSIM loss \cite{zhao2016loss} is used as a perceptual loss to maximize the structural similarity between both real and synthetic images. The SSIM can efficiently measure the perceptual difference between images with less sensitivity to luminance and contrast difference, and using it in training image translation models leads to better quality and more realistic synthetic images. The perceptual loss $	\mathcal{L}_p$ is defined as: 

\begin{equation}\label{Eqn_percept_loss}
	% \small
	\mathcal{L}_p = 1 - SSIM
\end{equation}

The overall MRI-to-PET translation loss, $\mathcal{L}$,  is defined as the weighted sum of both reconstruction and perceptual loss components:

\begin{equation}\label{Eqn_translation_loss}
	% \small
	\mathcal{L} = \lambda_{r}  \mathcal{L}_{r} + \lambda_{p}  \mathcal{L}_{p} 
\end{equation}
where $\lambda_r$ and $\lambda_p$ are the weights for reconstrtuction and perceptual loss terms, respectively. The proposed weighting is 0.2 for $\lambda_r$ and 0.8 for $\lambda_p$ based on grid search in hyperparameter optimization, which leads to prediction results that well-mimic subjective ratings. 

In our experiments, the Nesterov Adam optimizer \cite{dozat2016incorporating}, an improved variant of the Adam optimization algorithm \cite{kingma2014adam}, was used with a learning rate of 0.0002 and a batch size of 4. The proposed encoder-decoder network was trained with the proposed MRI-to-PET translation loss function for 150 epochs and early stopping of 20 epochs.

\subsection{Statistical Analyses}
% \subsection{Statistical Significance}
For both healthy controls and cerebrovascular disease patients, CBF was measured in 10 brain territories (anterior, 3 middle, and posterior in each of the right and left hemisphere) based roughly on the Alberta Stroke Program Early CT Score (ASPECTS, see Figure~\ref{fig.aspects}) \cite{barber2000validity}. Bland-Altman analyses were conducted to examine the regional CBF agreement between the true $^{15}$O-water PET CBF maps and each of the (i)~synthesized PET CBF maps obtained by our model, (ii)~SD-CBF maps measured with SD-ASL, and (iii)~and MD-CBF map measured with MD-ASL. Joint intensity scatter plots were also used to study the correlation between the regional CBF measurements before and after acetazolamide administration. The Pearson's correlation coefficient (r) was used to measure the strength of the linear association between the true $^{15}$O-water PET CBF and each of synthetic PET CBF, SD-CBF, and MD-CBF. Please note that the SD-CBF and MD-CBF are directly available from MRI, and any improvement from their prediction represents the added value of the trained network.

\subsection{Identifying Regional CBF Abnormalities}

CBF measured in the 10 vascular territories was used to examine regional CBF abnormalities and identify affected brain areas in the cerebrovascular disease patients. Since the global and regional values of CBF increase substantially after acetazolamide (Diamox) administration, different threshold CBF values are used to detect CBF abnormalities in Pre-Diamox and Post-Diamox CBF scans. In other words, the threshold CBF used to characterize the unusual vascular territories in the CBF maps acquired after acetazolamide administration was relatively higher than that used to characterize CBF maps acquired at baseline. In both conditions, threshold values were defined as 3 standard deviations below the mean PET CBF values in the healthy control participants. The receiver operating characteristic (ROC) curves were used to show the classification performance of synthetic PET CBF, SD-CBF, and MD-CBF at different discrimination thresholds before and after acetazolamide administration. The area under the ROC curve (AUC) scores were also measured to evaluate the diagnostic ability of synthetic PET CBF, SD-CBF, and MD-CBF to identify the vascular territories with abnormally low CBF.

%%%%%%%%%%%%%%%%%%%%%%%%%%%%%%%%%
\section{Experiments and Results}
\label{sec4}

We first evaluate the quantitative and qualitative performance of the proposed MRI-to-PET translation approach and demonstrate how synthetic images can support clinical decisions by improving the diagnosis and assessment of neurological conditions. Then, we perform ablation experiments to evaluate the effectiveness and contributions of different loss functions and the role of attention mechanisms. Finally, we study the relative impact of different input MRI scans on the overall quality of synthetic PET images.

\subsection{Experimental Setup}

In this retrospective cohort study, we analyzed simultaneous MRI/PET data from healthy subjects and cerebrovascular disease patients acquired from two cohorts. The participants of the first cohort (30 HCs and 40 PTs) had only one visit, at which time three PET scans were acquired from each participant (two before and one after ACZ administration). Four cases were excluded (1 HC and 3 PTs) because of missing either one of the MRI exams or the $^{15}$O-water PET scan. In the second cohort, the majority of the healthy control participants had two visits while all patients had only one visit to the imaging center. In each visit, two PET scans were acquired from each participant, one before and one after the ACZ administration. MRI examinations for each cohort included T1-w, T2-FLAIR, SD-ASL, MD-ASL, PD, ATT, SD-CBF, and MD-CBF. 

Figure~\ref{fig.setup} shows the analysis workflow for model training, validation, and independent testing. The database of both cohorts was divided into two separate sets. The first set included simultaneous PET/MRI data from 50 HCs and 45 PTs and it is used for model development and primary analysis. Fivefold cross-validation was used to examine the model performance on different portions of the data, where the data set is divided into five sub-groups: 3 were used for model training, 1 fold for validation and the remaining one for testing. To avoid data leakage, we were careful to include all data form the same subject (both baseline and post-acetazolamide) in any of the training, validation, and testing sets. To examine the ability of the trained model to reproduce results when deployed in practice, the second set of data, named ``Generalization Set'' was used to examine the model performance on unseen data acquired from 15 different participants (8 HCs and 7 PTs). A set of paired comparison methods were used to study the relationship between the regional CBF values of actual and synthetic PET scans, and whether synthetic CBF maps can be used to identify the regional CBF abnormalities.

\subsection{Quantitative Results}

The quantitative performance of our model was evaluated using NRMSE, PSNR, and SSIM. Theoretically, lower NRMSE values and higher PSNR and SSIM values indicate better quality of the synthetic images. Table~\ref{tab1} reports the average quantitative results for both healthy controls and cerebrovascular disease patients among the test sets. All assessment metrics were computed based on the whole three-dimensional brain region. Experiments clearly revealed that the PET synthesis results were markedly better for healthy controls than patients in terms of all metrics. The reason behind this performance disparity could be the structural abnormalities related to vascular disease in the brain and their different appearance among different MRI contrasts. The average results, also shown in Table~\ref{tab1}, demonstrate that our optimized encoder-decoder network can efficiently integrate multiple MRI exams and produce high-quality synthetic PET images.

In order to rigorously evaluate the predictive power of the proposed encoder-decoder network (``ours''), we compare its performance with other state-of-the-art image synthesis networks on the same dataset, including 2D~U-Net, 2D~conditional GAN (cGAN), 3D~U-Net, and 3D~cGAN. The results of the 2D~U-Net are generated using the same network implementation with the same parameters as in \cite{2020guo}. The 3D~U-Net had a similar implementation to our network but without specific modification or addition to the network elements. The 2D~cGAN and 3D~cGAN were implemented similar to \cite{wang20183d}. The quantitative evaluation of these networks as well as our encoder-decoder network is presented in Table~\ref{tab2}. The 2D~cGAN achieves on-par or slightly better performance than the 2D~U-Net with a statistically insignificant difference in the quality assessment metrics. Results also show that 3D~U-Net and 3D~cGAN produce comparable PET synthesis results, suggesting that simple encoder-decoder networks would be more practical than unstable GANs for this particular medical image generation problem. 

Further, the quantitative measures in Table~\ref{tab2} demonstrate that the 3D implementation of U-Net and cGAN outperforms their 2D counterparts, due to the increased level of spatial information in 3D images. Our attention-based encoder-decoder network with the custom loss function yields significantly better performance than competing methods across all performance metrics, which is consistent with our visual assessment. The performance of our PET synthesis approach demonstrates its ability to leverage the structural and contextual information in multi-contrast MRI data, thus improving the generated PET image quality.

\begin{table}[!t]
	\caption{\label{tab1} PET synthesis results for healthy controls and cerebrovascular disease patients. The model is evaluated using five-fold cross-validation, and the quantitative metrics are computed for the whole brain region. $\uparrow$/$\downarrow$~denotes that higher/lower values correspond to better quality of synthetic PET. Results are presented as mean $\pm$ standard deviation (std).}  
	\small
	\centering
	\begin{tabular}{|c|c|c|c|}
		\hline
		Participants & NRMSE $ \downarrow $ & PSNR (dB) $ \uparrow $ & SSIM $ \uparrow $\\
		& mean $\pm$ std & mean $\pm$ std & mean $\pm$ std \\
		\hline
		Controls & 0.024 $\pm$ 0.010 & 40.16 $\pm$ 1.09& 0.940 $\pm$ 0.008 \\
		\hline
		Patients  & 0.062 $\pm$ 0.015 & 37.32 $\pm$ 1.17 & 0.912 $\pm$ 0.012 \\
		\hline
		\hline
		Average        & 0.044 $\pm$ 0.015 & 38.80 $\pm$ 1.18 & 0.924 $\pm$ 0.014 \\
		\hline
		
	\end{tabular}
\end{table}

\begin{table}[!t]
	\caption{\label{tab2} Quantitative comparison between our model and baseline models. Statistically significant results are highlighted in \textbf{bold} font.}		
	\small
	\centering
	\begin{tabular}{|c|c|c|c|}
		\hline
		Mathod & NRMSE $ \downarrow $ & PSNR (dB) $ \uparrow $ & SSIM $ \uparrow $\\
		& mean $\pm$ std & mean $\pm$ std & mean $\pm$ std \\
		\hline
		% U-Net (2D) & 0.209 $\pm$ 0.039 & 29.45 $\pm$ 1.92 & 0.854 $\pm$ 0.036 \\
		U-Net (2D) & 0.204 $\pm$ 0.034 & 30.45 $\pm$ 1.92 & 0.862 $\pm$ 0.030 \\
		\hline
		cGAN (2D) & 0.205 $\pm$ 0.032 & 31.70 $\pm$ 2.04 & 0.865 $\pm$ 0.028 \\
		\hline
		U-Net (3D) & 0.168 $\pm$ 0.024 & 33.86 $\pm$ 1.35 & 0.880 $\pm$ 0.015 \\
		\hline
		cGAN (3D) & 0.168 $\pm$ 0.026 & 34.00 $\pm$ 1.42 & 0.878 $\pm$ 0.016 \\
		\hline
		\textbf{Ours }          & \textbf{0.044} $\pm$ \textbf{0.015} & \textbf{38.80} $\pm$ \textbf{1.18} & \textbf{0.924} $\pm$ \textbf{0.014} \\
		\hline		
	\end{tabular}
\end{table}

\begin{figure*}[!ht]
	\centering
	\includegraphics[width=\textwidth]{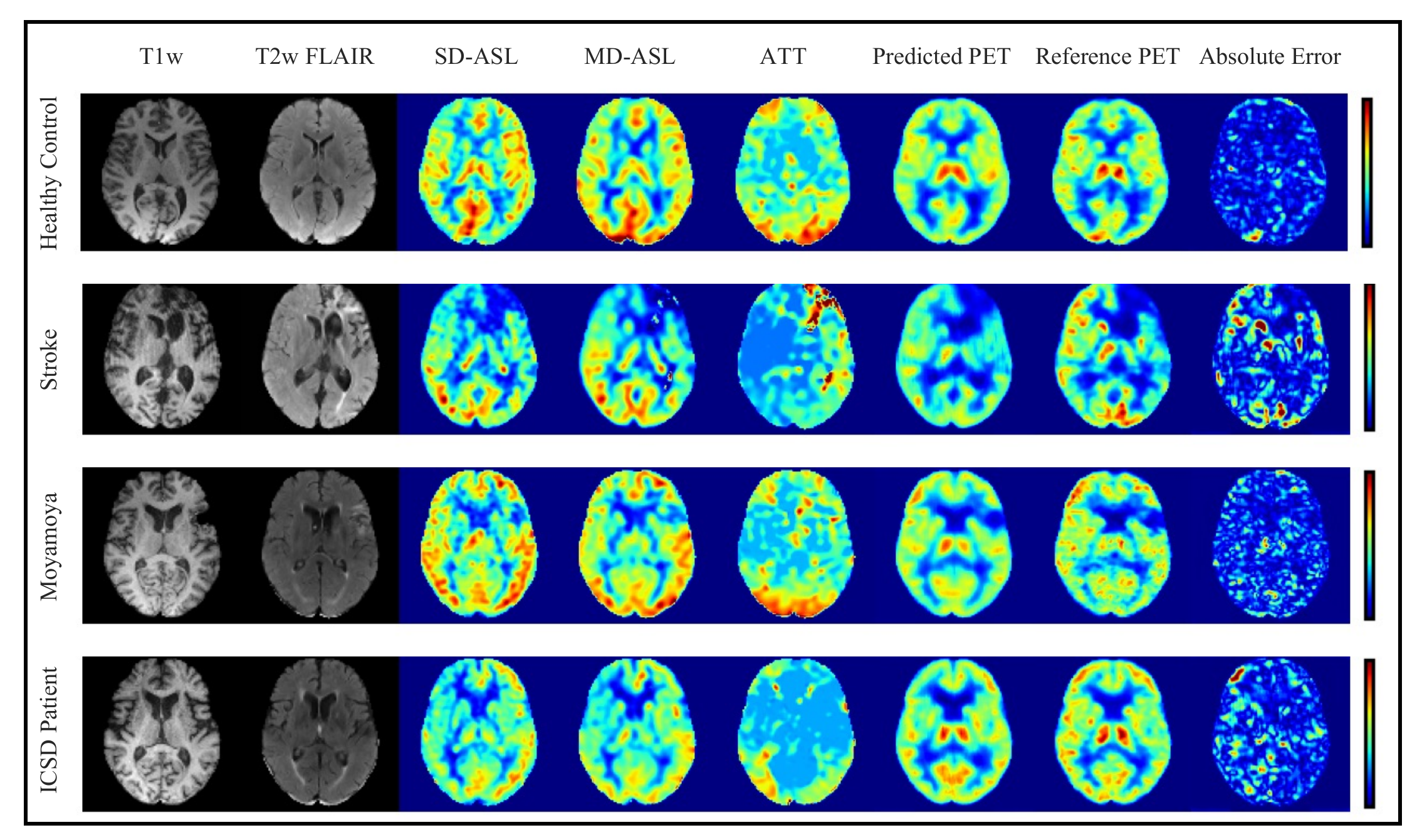}
	\caption{MRI-to-PET prediction results for healthy control and cerebrovascular disease patients in the axial plane: Examples of input multi-contrast MRI scans, output synthetic PET, reference PET, and corresponding magnified ($\times$3) absolute error maps.}
	\label{fig.axialprediction}
\end{figure*}

\subsection{Qualitative Results}

We demonstrate the significance of leveraging both structural and perceptual information in multi-contrast MRI data through optimizing the custom loss function and the use of attention mechanisms. To this end, we conduct a qualitative comparison between the synthetic and actual PET images for representative healthy controls and cerebrovascular disease patients. Figure~\ref{fig.axialprediction} shows examples of input MRI scans, synthesized PET images obtained by the model, reference PET images, and corresponding magnified Absolute Error maps of representative test samples. The first row reveals the model qualitative performance for a normal subject without abnormal brain lesions, where the generated PET image was indistinguishable from the ground-truth PET image. The remaining three rows show the model performance for patients with ischemic stroke, Moyamoya disease, and intracranial atherosclerotic steno-occlusive disease (ICSD). 

\begin{figure*}[!ht]
	\centering
	\includegraphics[width=\textwidth]{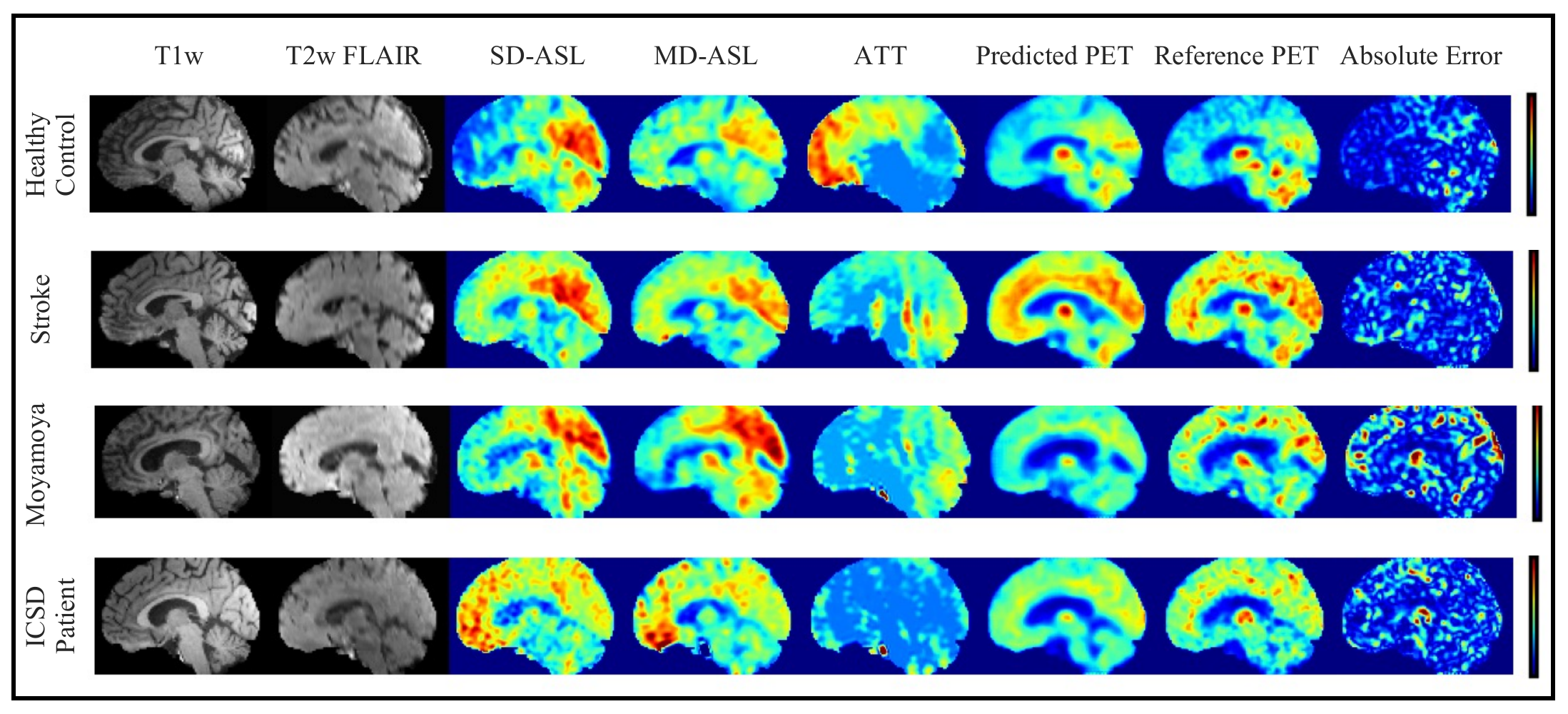}
	\caption{MRI-to-PET prediction results for healthy control and cerebrovascular disease patients in the sagittal plane: Examples of input multi-contrast MRI scans, output synthetic PET, reference PET, and corresponding magnified ($\times$3) absolute error maps.}
	\label{fig.sagittalprediction}
\end{figure*}

\begin{figure*}[!ht]
	\centering
	\includegraphics[width=\textwidth]{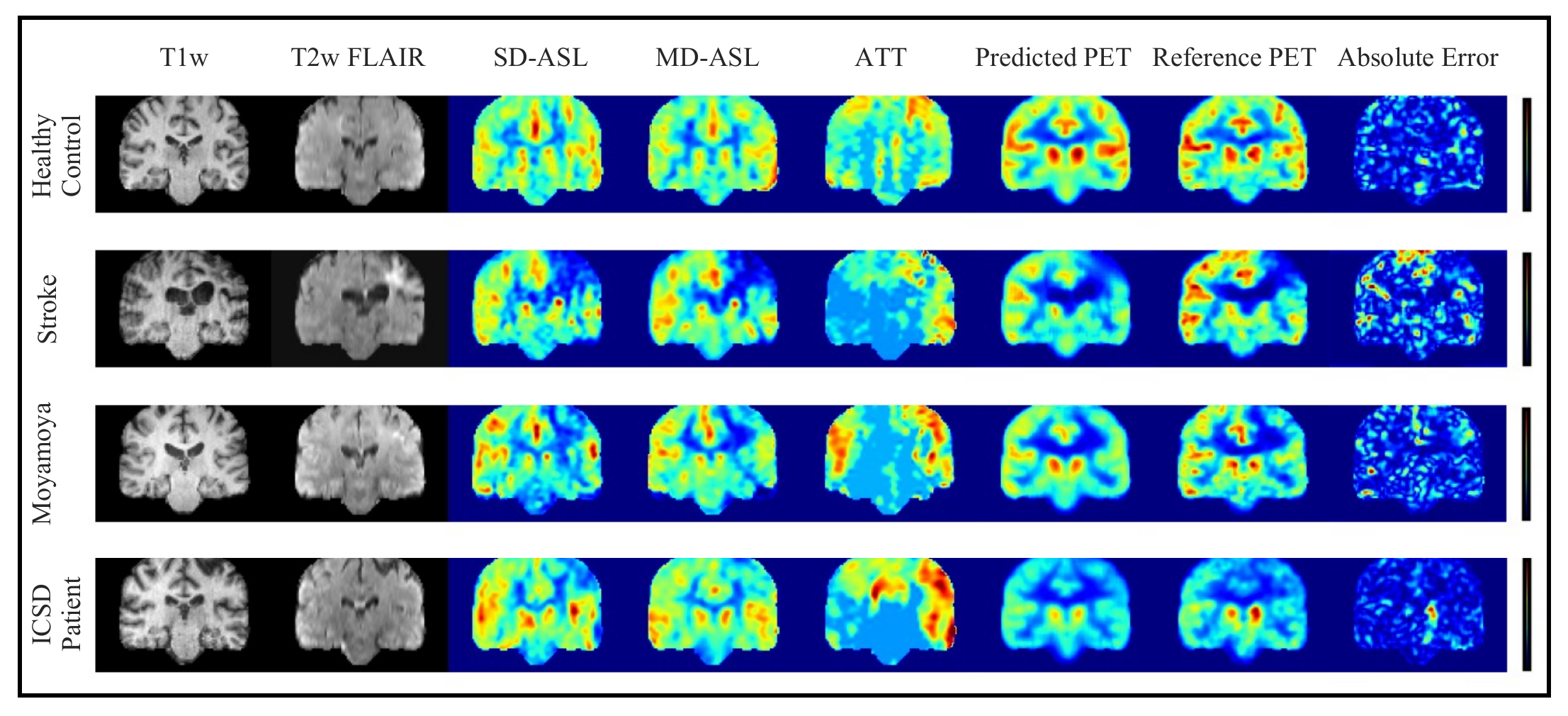}
	\caption{MRI-to-PET prediction results for healthy control and cerebrovascular disease patients in the coronal plane: Examples of input multi-contrast MRI scans, output synthetic PET, reference PET, and corresponding magnified ($\times$3) absolute error maps.}
	\label{fig.coronalprediction}
\end{figure*}

Figures~\ref{fig.sagittalprediction} and \ref{fig.coronalprediction} show the qualitative visualizations of synthetic PET scans for the same representative subjects along the sagittal and coronal planes, respectively. They reveal an outstanding PET prediction performance for the healthy controls in both sagittal and coronal views of the brain. For cerebrovascular patients with impaired regional CBF, the models are more inclined to overfit to normal regions since the abnormal brain regions are limited relative to the whole brain volume. Therefore, the prediction performance for areas with altered CBF (\textit{e.g.}, ischemic penumbra and infarcted tissue) is somewhat inferior to that in the normal brain areas. Nevertheless, our model often produces superior visual performance for the abnormal regions when compared to the baseline image synthesis networks, which miss or overestimate the CBF values in abnormal brain territories. Overall, our PET synthesis approach can effectively improve the quality and clinical utility of the structural and perfusion MRI exams, producing high-quality synthetic $^{15}$O-water PET CBF maps without using radioactive tracers. 

\subsection{Ablation Study}

Ablations studies are a valuable method to investigate knowledge representations in encoder-decoder networks and are especially helpful to examine network performance and reliability against structural artifacts and ghosting. We performed experiments to evaluate the effectiveness and contributions of different network settings and training strategies. We have experimented with several loss functions including MSE, MAE, and SSIM, as well as a custom loss function with weighted summation of different metrics, to optimize the quality of synthetic PET CBF maps. We further studied the importance of attention mechanisms used by the network’s decoder for the PET CBF prediction solely based on a subset of the encoder's feature maps. To be precise, in this section, we do not only remove parts of the network but also substitute them with more appropriate alternative constructs. 

\begin{figure*}[!ht]
	\centering
	\includegraphics[width=\textwidth]{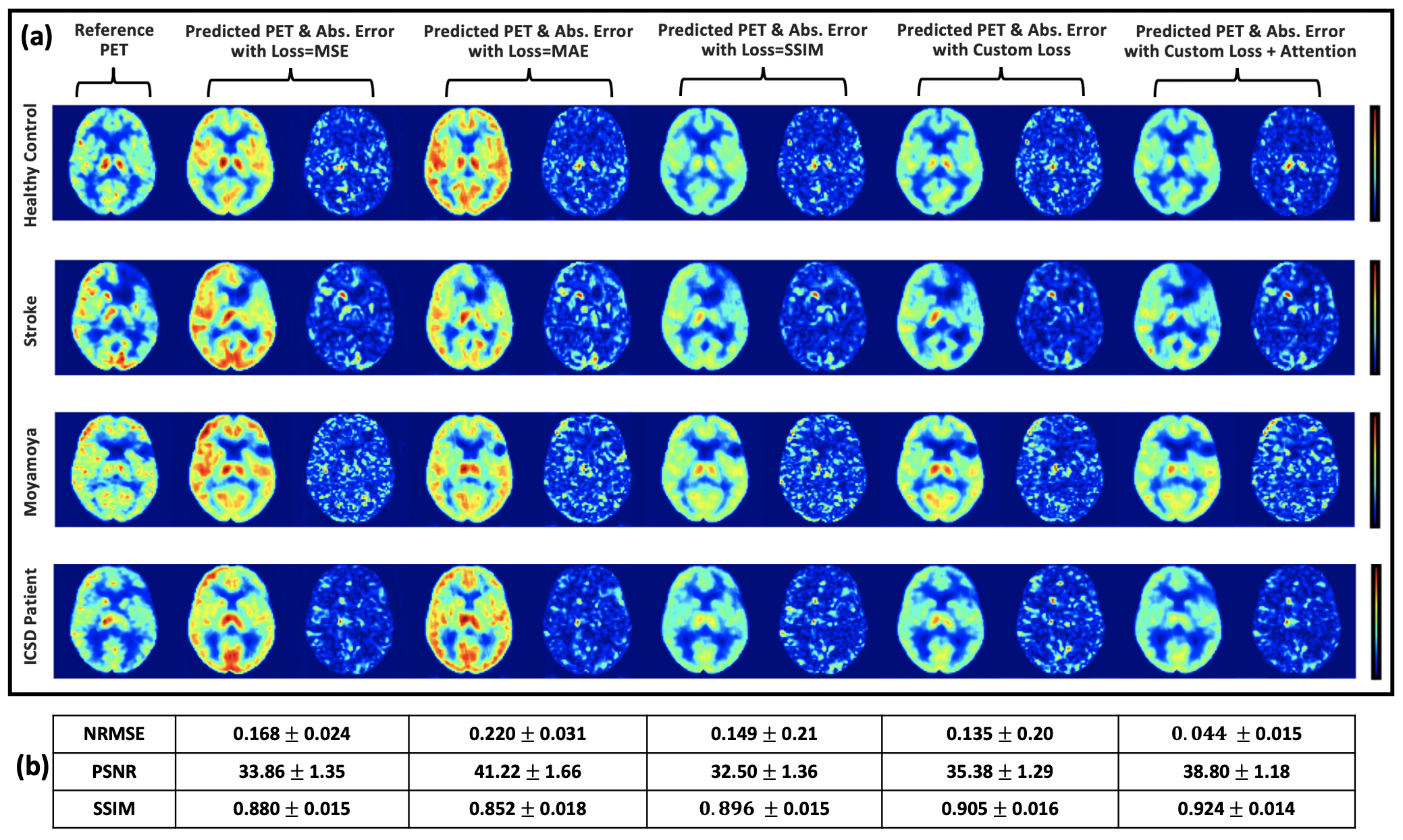}
	\caption{Examples of PET CBF prediction for healthy controls and cerebrovascular disease patients at different loss functions and network settings: (a)~example results of the reference PET CBF against synthetic PET CBF and magnified ($\times$3) absolute error maps at different settings in the axial plane, (b)~quantitative comparison between different loss functions and network elements.}
	\label{fig.ablationstudy}
\end{figure*}

Figure~\ref{fig.ablationstudy} shows the PET prediction performance of different loss functions and network settings and the incremental performance gain of each component. The reference PET, synthetic PET, and also the corresponding absolute error map produced at different network settings are displayed for both healthy subjects and patients with cerebrovascular diseases (Figure~\ref{fig.ablationstudy} (a)). Steady improvement in both quantitative and qualitative CBF prediction results was observed as we used more appropriate loss functions (Figure~\ref{fig.ablationstudy} (b)). Also, attention mechanisms were found to have a crucial role in enabling the network to focus more on the relevant aspects of the input at the channel and spatial levels, thus yielding improved synthetic PET CBF quality. 

The quantitative results also showed that an attention-based encoder-decoder network with a custom loss function produced superior PET CBF prediction results, achieving SSIM of 0.924, NRMSE of 0.044, and PSNR of 38.80 dB. On the whole, 3D encoder-decoder networks along with attention mechanisms and our custom loss function were able to effectively exploit channel-level and spatial-level information to produce high-quality PET CBF maps.

\subsection{CBF Quantification Assessment}

The statistical significance of experimental results was evaluated using a set of paired comparison analyses. The generalization set was used to examine the levels of agreement and correlation between regional CBF values in true ${15}$O-water PET CBF measurements and those of synthetic PET CBF, SD-CBF, and MD-CBF maps. The generalization set included 60 simultaneous PET/MRI scans acquired from 10 healthy controls and 11 cerebrovascular disease patients. The 10 control cases had a total of 32 PET/MRI scans (18 pre-acetazolamide and 14 post-acetazolamide) and the 11 patients had a total of 28 PET/MRI scans (17 pre-acetazolamide and 11 post-acetazolamide). Overall, 600 vascular territories from the 60 PET/MRI observations were used for the paired comparison analyses between synthesized PET CBF obtained by our encoder-decoder network and MRI-derived CBF measurements. 

\begin{figure*}[!ht]
	\centering
	\includegraphics[width=\textwidth]{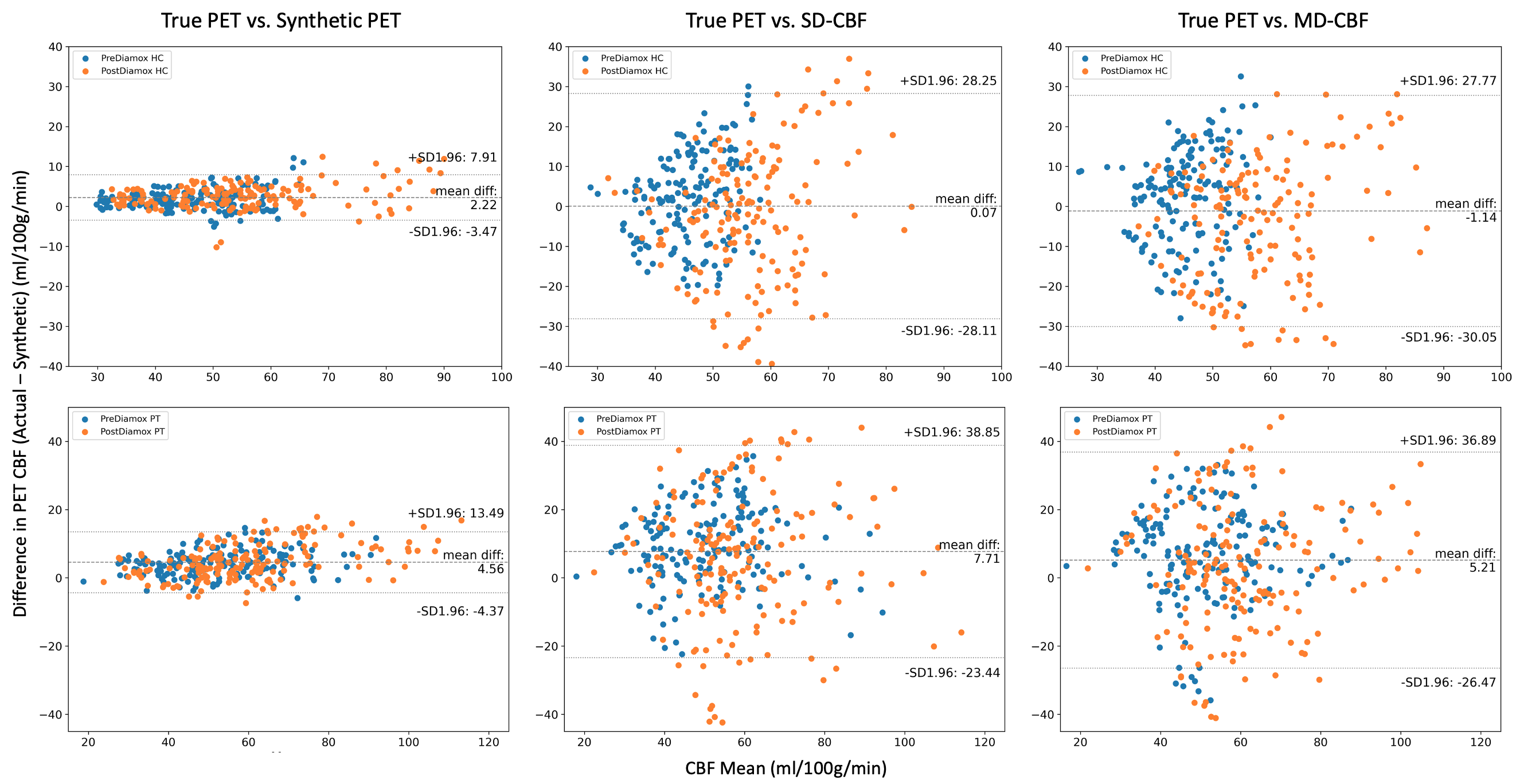}
	\caption{Bland-Altman plots of mean CBF in ASPECTS vascular territories for healthy control (HC, top panel) and cerebrovascular disease patients (PT, bottom panel): Each panel includes three plots showing the agreement between the reference PET CBF (True PET) and (i)~the PET CBF produced by the encoder-decoder network (Synthetic PET, left), (ii)~CBF quantified from single-delay ASL (SD-CBF, middle), and (iii)~CBF quantified from multi-delay ASL (MD-CBF, right).}
	\label{fig.blandaltman}
\end{figure*}

Figure~\ref{fig.blandaltman} displays the Bland-Altman plots of regional CBF values for healthy control (top panel) and cerebrovascular disease patients (bottom panel). Each panel includes three plots to illustrate the levels of agreement between the reference PET CBF maps and each of the synthetic PET, SD-CB, and MD-CBF maps. In healthy controls, the regional CBF values of synthetic PET showed comparable bias and significantly lower variance when compared to the two ASL-derived CBF maps, demonstrating the added value of the trained network. The synthetic PET CBF maps produced by our encoder-decoder network were only 2.2 ml/100g/min lower than true PET CBF maps with 95\% confidence intervals of -3.5 \& 7.9 ml/100g/min. In cerebrovascular disease patients, synthetic PET CBF maps also demonstrated less bias and markedly smaller variance than ASL-derived CBF maps. Moreover, the PET synthesis performance in patients was slightly inferior to that in healthy control participants, showing an average bias of 4.6 ml/100g/min with 95\% confidence intervals of -4.4 \& 13.5 ml/100g/min. 

\begin{table*}[!t]
	\caption{\label{tab3} PET synthesis error (true PET CBF $-$ synthetic PET CBF): Bias and precision in synthetic PET measurements for healthy controls and patients at different scan times. Mean and SD represent the bias and variability in the measurements, respectively. No. refers to the number of PET/MRI observations.}		
	\centering
	\begin{tabular}{|c|ccc|ccc|ccc|}
		\hline
		Group     & \multicolumn{3}{c|}{Controls}                                            & \multicolumn{3}{c|}{Patients}                                            & \multicolumn{3}{c|}{Average}                                             \\ \hline
		Scan Time & \multicolumn{1}{c|}{PreDiamox} & \multicolumn{1}{c|}{PostDiamox} & Total & \multicolumn{1}{c|}{PreDiamox} & \multicolumn{1}{c|}{PostDiamox} & Total & \multicolumn{1}{c|}{PreDiamox} & \multicolumn{1}{c|}{PostDiamox} & Total \\ \hline
		Mean      & \multicolumn{1}{c|}{1.82}      & \multicolumn{1}{c|}{2.62}       & 2.22  & \multicolumn{1}{c|}{4.25}      & \multicolumn{1}{c|}{4.88}       & 4.56  & \multicolumn{1}{c|}{3.05}      & \multicolumn{1}{c|}{3.80}       & 3.42  \\ \hline
		SD        & \multicolumn{1}{c|}{2.56}      & \multicolumn{1}{c|}{3.20}       & 2.90  & \multicolumn{1}{c|}{3.68}      & \multicolumn{1}{c|}{5.25}       & 4.56  & \multicolumn{1}{c|}{3.40}      & \multicolumn{1}{c|}{4.54}       & 4.04  \\ \hline
		No.         & \multicolumn{1}{c|}{180}       & \multicolumn{1}{c|}{140}        & 320   & \multicolumn{1}{c|}{170}       & \multicolumn{1}{c|}{110}        & 280   & \multicolumn{1}{c|}{350}       & \multicolumn{1}{c|}{250}        & 600   \\ \hline
	\end{tabular}
\end{table*}

Bland-Altman plots also demonstrate that the mean CBF was not significantly different among CBF measurement types (p=0.14), but variability in SD-CBF and MD-CBF measurements was significantly higher than synthetic PET CBF measurements (p$<$0.0001). Table~\ref{tab3} reports the detailed PET synthesis error, showing the bias and variability in synthetic PET CBF measurements for healthy controls and patients before and after acetazolamide administration. It can be noticed that both bias and precision in synthetic PET measurements differed between groups and between timepoints (all four marginal comparisons with p$<$0.001). Specifically, regional CBF values in healthy controls showed significantly lower bias and precision than in patients for both baseline and post-acetazolamide timepoints. The PET synthesis performance was also found to be moderately better at baseline conditions.

\begin{figure*}[!ht]
	\centering
	\includegraphics[width=\textwidth]{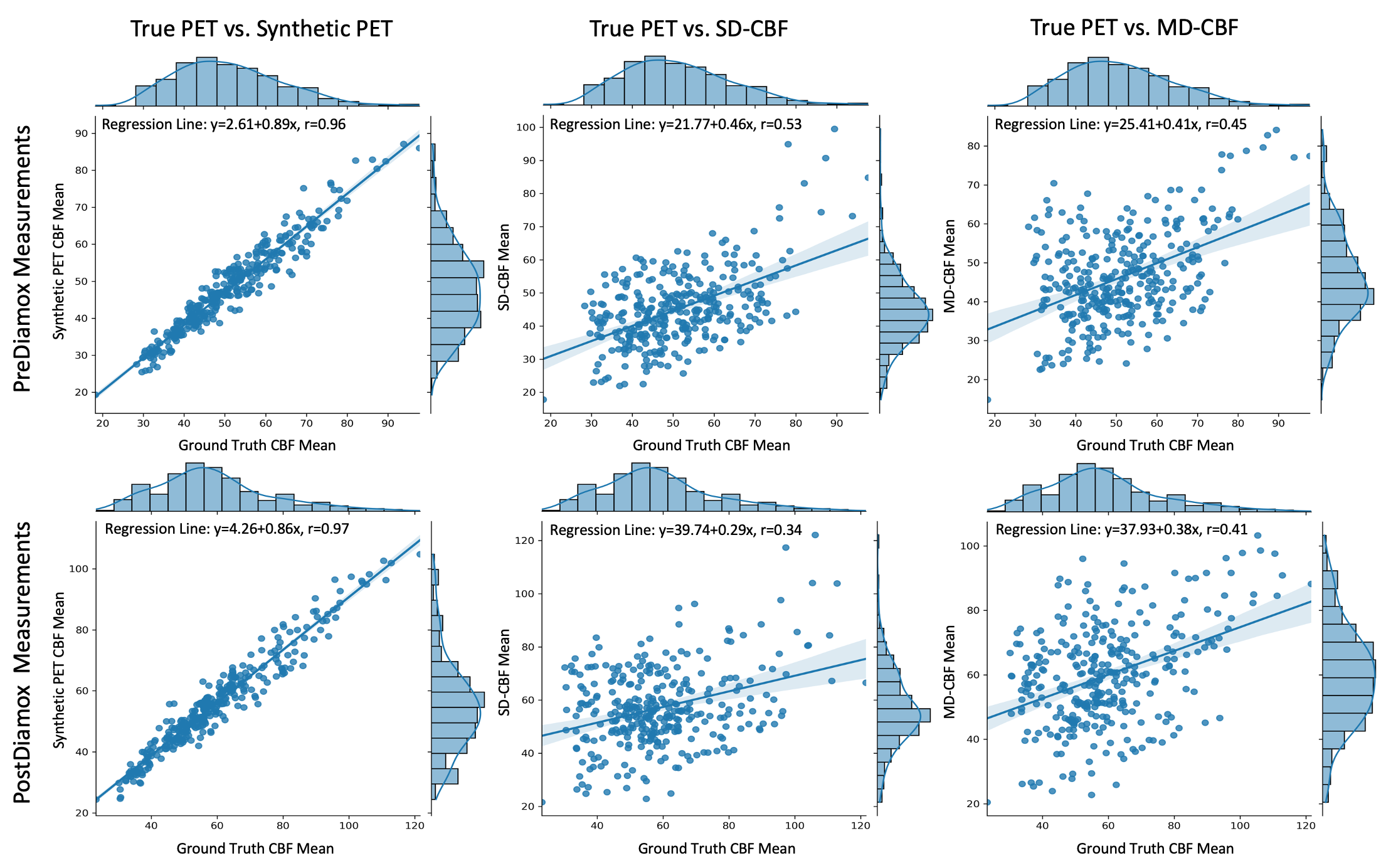} %.97
	\caption{Joint plots of mean CBF in ASPECTS vascular territories for PreDiamox (top panel) and PostDiamox (bottom panel) measurements: Each panel displays three plots for the relationship and distribution histogram between True PET and Synthetic PET~(left), SD-CBF~(middle), and MD-CBF~(right). The regression line and Pearson correlation coefficient (r) are added to each of the joint plots.}
	\label{fig.jointplot}
\end{figure*}

The correlation between regional CBF measurements in acquired $^{15}$O-water PET and those of synthetic PET and ASL-derived CBF maps was also investigated in this study. Figure~\ref{fig.jointplot} describes the density and joint scatter plots of regional CBF values in pre-acetazolamide (top panel) and post-acetazolamide (bottom panel) measurements. Each of the top and bottom panels in Figure~\ref{fig.jointplot} displays three plots for the relationship and distribution histogram between true $^{15}$O-water PET CBF and synthetic PET CBF~(left), SD-CBF (middle), and MD-CBF (right). The regression line and Pearson's correlation coefficient (r) are added to each of the joint plots. From the plots, we can see a high positive correlation between true and synthetic PET's regional CBF values. The Pearson's correlation coefficient was found to be 0.96 and 0.97 for the pre-acetazolamide and post-acetazolamide scans, respectively. The ASL-derived CBF maps, on the other hand, showed a moderate positive correlation (r=0.45-0.53) at baseline and a weak positive correlation (0.34-41) for post-acetazolamide measurements.

\subsection{Clinical Significance -- Abnormal Region Identification}

\begin{figure*}[!ht]
	\centering
	\includegraphics[width=\textwidth]{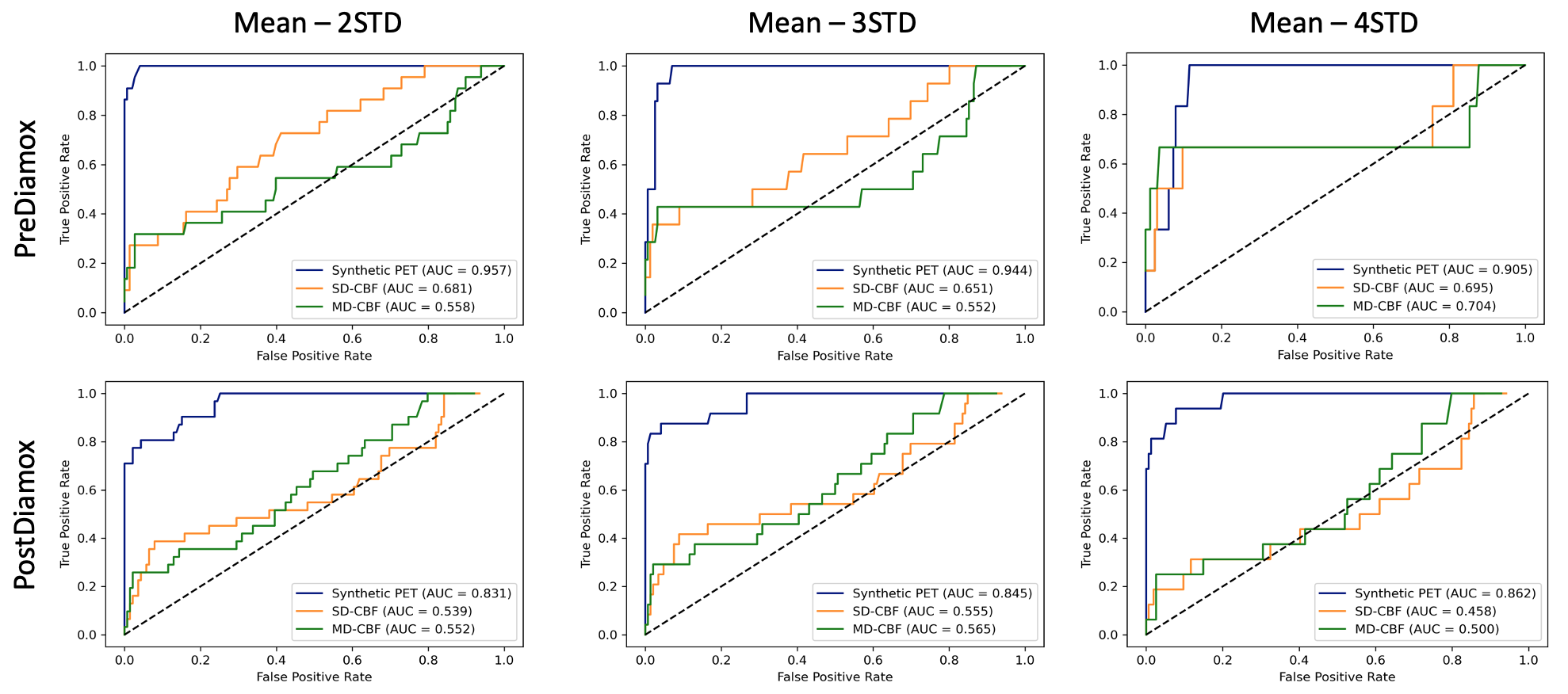}
	\caption{ROC curves and AUC scores for identifying vascular territories with reduced CBF in Prediamox (top panel) and PostDiamox (bottom panel) measurements for cerebrovascular patients: Each panel includes three plots showing the classification performance at three different threshold values, (i)~Threshold at 2~standard deviation (STD) below the mean CBF of healthy control participants (left), (ii)~Threshold at 3 STD below mean CBF (middle), and (iii)~Threshold at 4 STD below mean CBF (right). Each plot includes three ROC curves showing the classification performance of Synthetic PET (blue curve), SD-CBF (red curve), and MD-CBF (green curve).}
	\label{fig.auc}
\end{figure*}

To assess the feasibility and clinical value of the proposed MRI-to-PET translation method, the utility of synthetic PET images was tested for identifying regional CBF abnormalities in cerebrovascular disease patients. Different threshold CBF values were used to examine the diagnostic performance of both synthetic PET CBF and ASL-derived CBF maps. The tested thresholds were defined as 2, 3, and 4 standard deviations below the mean CBF values of these regions in healthy control participants. Figure~\ref{fig.auc} shows the classification performance of synthetic PET CBF, SD-CBF, and MD-CBF to identify reduced CBF regions in pre-acetazolamide and post-acetazolamide CBF measurements. The plots show the ROC curves and AUC scores to differentiate between the vascular territories with abnormally low CBF and those with normal CBF. The synthetic PET CBF maps generated by our model outperforms ASL-derived CBF maps at all tested thresholds before and after acetazolamide administration. 

\begin{figure*}[!ht]
	\centering
	\includegraphics[width=\textwidth]{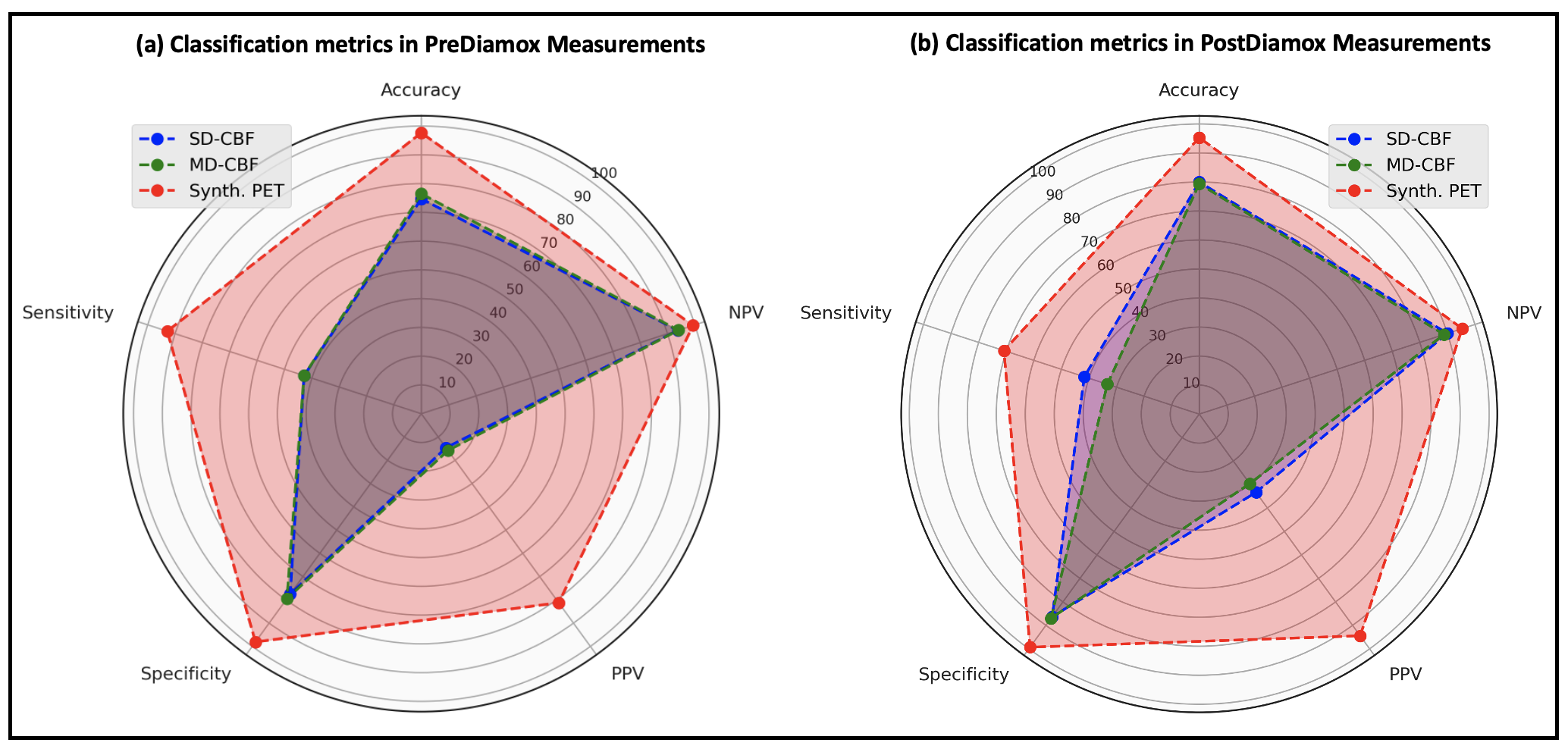}
	\caption{Radar charts of classification performance measures for SD-CBF, MD-CBF, and synthetic PET CBF at Threshold = Mean -- 3 STD: (a) and~(b) Evaluation metrics for detecting abnormal regions (\textit{i.e.}, regions with reduced CBF) in PreDiamox and PostDiamox measurements, respectively.}
	\label{fig.radar}
\end{figure*}

In pre-acetazolamide measurements, the synthetic PET CBF consistently achieved significantly higher classification accuracy (AUC=0.905-0.957) compared to SD-CBF (AUC=0.651-0.695) and MD-CBF (AUC=0.552-0.704) at the threshold CBF values of 2-4 standard deviation below the mean in healthy control participants. Likewise, synthetic PET CBF achieved superior classification performance to the ASL-derived CBF maps under the post-acetazolamide conditions. AUC scores between 0.831-0.862, 0.458-0.555, and 0.500-0.565 were achieved by synthetic PET CBF, SD-CBF, and MD-CBF maps, respectively. Figure~\ref{fig.radar} also shows radar charts of the classification performance measures for SD-CBF, MD-CBF, and synthetic PET CBF at the threshold of 3 standard deviation below the mean CBF. The metrics of classification accuracy, sensitivity, specificity, positive predictive value (PPV), and negative predictive value (NPV) were used to evaluate the detection performance of abnormal regions in pre-acetazolamide and post-acetazolamide CBF measurements. These results demonstrate the diagnostic value of synthetic PET and how the proposed model improved the clinical utility of MRI-derived CBF measurements at both baseline and after vasodilator administration. Our PET synthesis model offers a great promise for medical diagnostics, showing accurate identification of regional CBF abnormalities in patients with cerebrovascular diseases. 

\subsection{PET Synthesis from Perfusion MRI without Structural Information}

In this section, we investigate the PET synthesis performance of the proposed encoder-decoder network when provided with ASL MRI images only as inputs. Yousefi \textit{et al.} \cite{2021yousefi} have previously studied the ASL-to-PET translation problem, where they used residual CNNs to generate PET data from 2D ASL and T1w images. Using seven-fold cross-validation, their PET synthesis network achieved SSIM of 0.85$\pm$0.08 and PSNR of 21.8$\pm$4.5 for healthy control participants. In our experiments, we only used single-delay and multi-delay ASL images as the only input to our encoder-decoder network while relaxing the need for anatomical information from any of the structural MRI scans. Our method shows better quantitative results, achieving an average SSIM and PSNR of 0.86$\pm$0.03 and 30.4$\pm$2.3, respectively. 

\begin{figure*}[!ht]
	\centering
	\includegraphics[width=\textwidth]{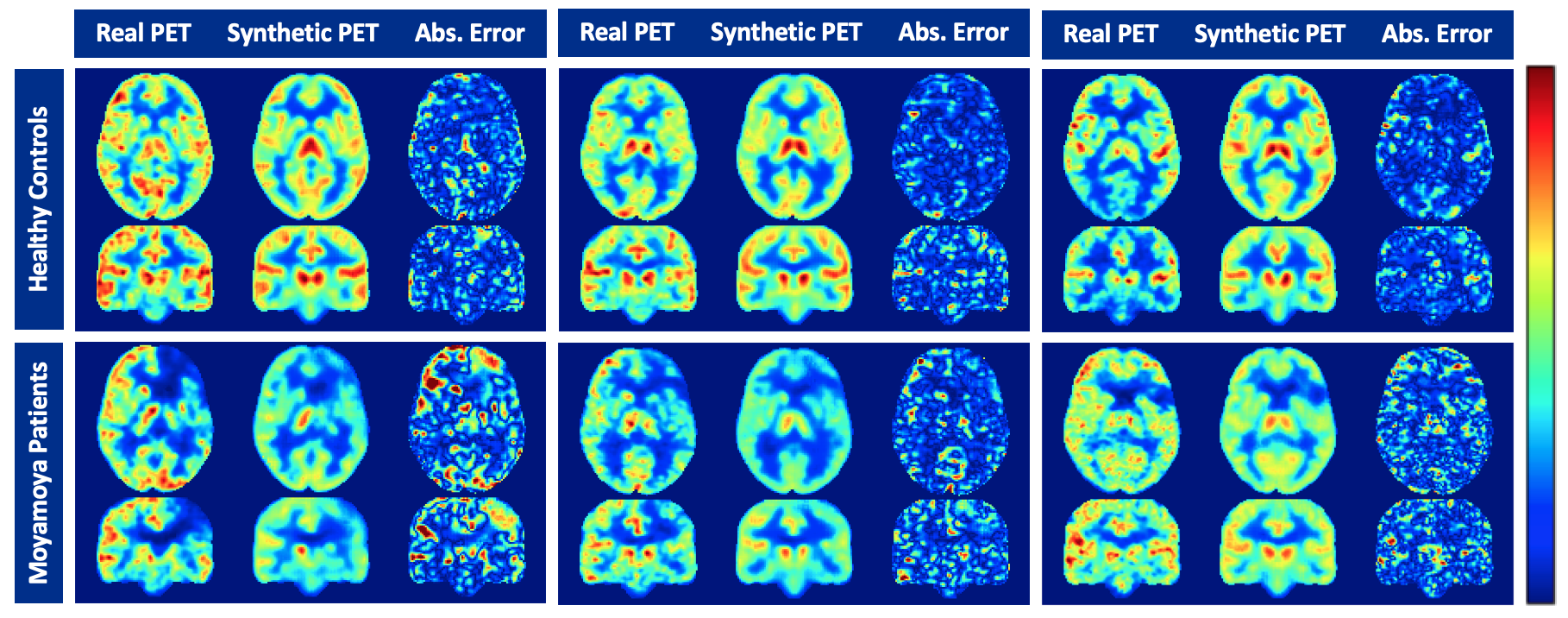}
	\caption{MRI-to-PET CBF prediction using only ASL MRI without structural imaging: Synthetic versus real PET axial and coronal brain images of 3 healthy controls (top row) and 3 Moyamoya disease patients (bottom row). Each panel represents a separate subject with real axial and coronal images and corresponding synthetic images and magnified ($\times$3) absolute error maps.}
	\label{fig.prediction_asl}
\end{figure*}

Figure~\ref{fig.prediction_asl} shows examples of ASL-to-PET prediction for both healthy controls and cerebrovascular disease patients in axial and coronal planes. It can be observed that our model produced adequate PET CBF maps for healthy controls, in which the magnified absolute error maps show an insignificant difference between the true and synthetic PET CBF maps. However, an inferior PET synthesis performance was seen in patients, showing overestimation for the brain regions with reduced CBF. In normal brain territories, the CBF was either underestimated or overestimated, showing a non-trivial discrepancy between true and synthetic PET images. The absence of anatomical structure (from T1w or T2-FLAIR) is probably the reason behind the performance deterioration.

\subsection{PET Synthesis from Structural MRI without ASL Imaging}

\begin{figure*}[!ht]
	\centering
	\includegraphics[width=\textwidth]{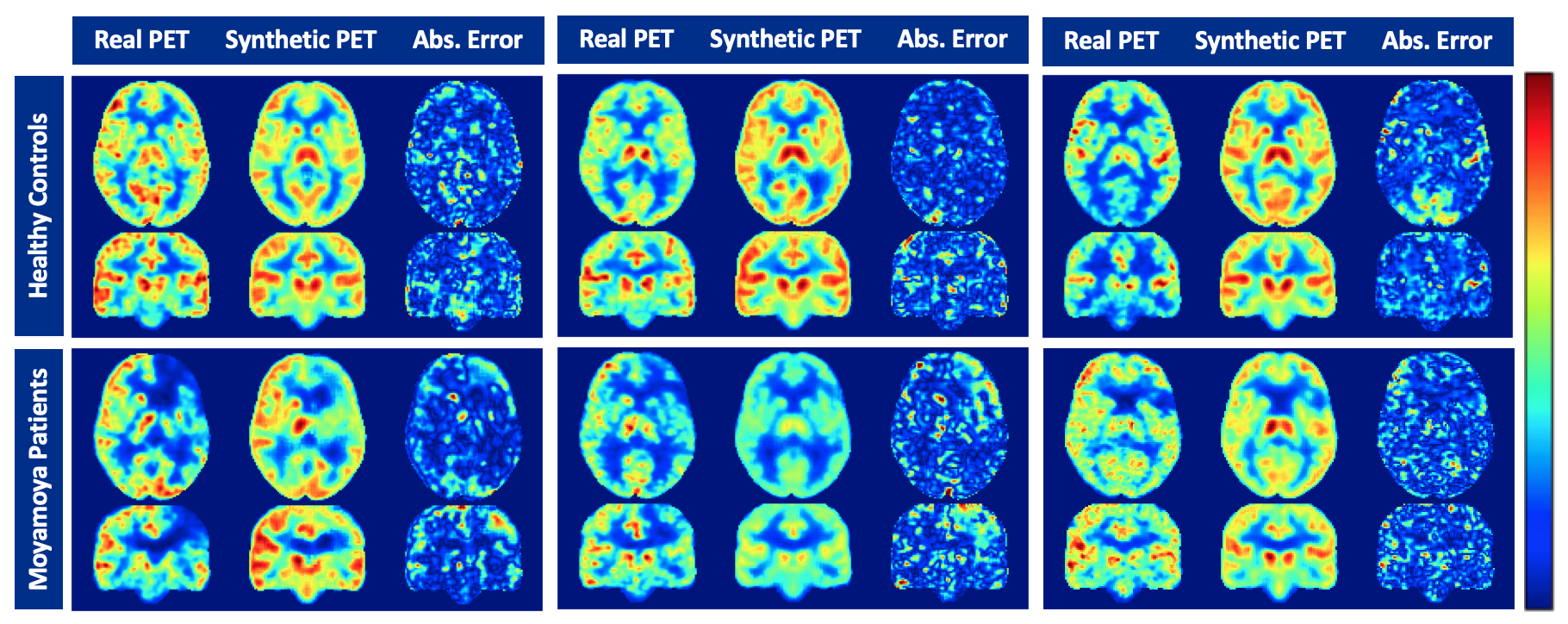}
	\caption{MRI-to-PET CBF prediction using only structural sequences (T1-weighted and T2-FLAIR) without ASL perfusion imaging: Synthetic versus real PET axial and coronal brain images of 3 healthy controls (top row) and 3 Moyamoya disease patients (bottom row). Each panel represents a separate subject with real axial and coronal images and corresponding synthetic images and magnified ($\times$3) absolute error maps.}
	\label{fig.prediction_struct_mri}
\end{figure*}

Lastly, we evaluated the feasibility of synthesizing PET scans from structural MRI including T1w and T2-FLAIR exams, but excluding data from the perfusion imaging sequences. Figure~\ref{fig.prediction_struct_mri} shows the results of structural MRI-to-PET translation for healthy controls and cerebrovascular disease patients. Our model showed moderate synthesis performance in healthy control cases, with a reasonable similarity between true and synthetic PET CBF maps. The whole-brain region was somewhat overestimated but the discrepancy between true and synthetic PET images was minor. This is not surprising performance, since in normal subjects, gray and white matter perfusion differ in a reproducible way.

On the other hand, a serious performance degradation was observed in PET image generation for cerebrovascular disease patients. Both axial and coronal visualizations reveal the limited performance and the inability of this model to produce acceptable CBF maps when provided with structural MRI as inputs. None of the brain regions with reduced CBF was properly predicted, showing clearly that ASL scans are crucial for PET image synthesis in patients with cerebrovascular disease.

\section{Discussion}

In this study, we presented and evaluated a 3D attention-based encoder-decoder network for brain MRI-to-PET translation. The proposed multimodal architecture effectively integrates structural MRI and ASL scans to capture both anatomical and perfusion features needed to improve the quality of synthesized PET scans. A custom loss function was developed to optimize the PET synthesis performance in both normal and abnormal brain regions. The proposed loss function is a combination of multiple loss components that work cooperatively on driving the network toward the most representative distribution of actual PET images. A reconstruction loss based on the mean absolute error was used to ensure high voxel-wise similarity between real and synthetic PET images. A perceptual loss based on SSIM was also used to supplement the global loss and maximize the contextual and visual similarity between real and synthetic PET images. Lastly, attention mechanisms were incorporated to capture long-range feature interactions and help the encoder-decoder network reliably learn the underlying multimodal data distribution. Results demonstrate that 3D convolutional encoder-decoder networks with attention mechanisms and a well-designed loss function can accurately synthesize PET CBF maps from multi-contrast MRI images without using radioactive tracers. 

We would like to emphasize the value of a network that can take a widely available modality (such as MRI) and predict results that are only available at extremely specialized centers ($^{15}$O-water PET).  This has the potential to enable gold-standard CBF measurements in sites that do not have access to short half-life PET agents, enabling whole new classes of research experiments that can be performed.  It also democratizes PET by allowing imaging at sites that economically cannot support a PET scanner and the expensive infrastructure that it requires.  This will allow accurate studies of CBF in a wider range of patients and disease classes, rather than restricting them to patients from urban areas with chronic conditions.

Several ablation studies were performed to illustrate the impact of different loss functions, network elements, and input MRI contrasts on the quality of synthesized PET images. Experimental results showed that both anatomical and functional information in structural and perfusion MRI exams are crucial for synthesizing high-quality and realistic PET scans, especially for patients with cerebrovascular diseases. Single-delay and multi-delay ASL scans had the greatest impact on the PET synthesis accuracy. Further, pairwise comparison methods such as Bland-Altman analyses and density scatter plots show a high level of agreement and correlation between regional CBF values in actual and synthetic PET images. In comparison to ASL-derived CBF measurements, the synthetic PET CBF maps demonstrated comparable bias, considerably better precision, and markedly higher positive correlation with true $^{15}$O-water PET CBF measurements.

The proposed work has several potential clinical applications. In contrast to previous MRI-to-PET translation studies, we report results on a task of clinical importance, namely discriminating between vascular territories with and without CBF abnormalities. The mean CBF values were computed for 10 brain regions in healthy control participants before and after administration of acetazolamide, a short-acting vasodilator. Different threshold CBF values based on mean CBF and its variability in healthy controls were then used to identify abnormal regions with low CBF in cerebrovascular disease patients. The improved performance of the network over MRI-only imaging demonstrates that the network is effective not only at characterizing the overall pattern of CBF, but specifically can predict the severity and location of abnormal regions, which may only occupy a small fraction of the overall image volume. Such information is not always captured in summary statistics often used for quantitative assessment, such as PSNR, NRMSE, and SSIM.

Data curation was one of the major limitations in this study. The co-registration of input multi-contrast brain MRI images and quantification of $^{15}$O-water PET CBF maps are laborious and time-consuming procedures. Subjective evaluations are also needed to ensure acceptable image and associated information quality. The deployment of automated deep learning algorithms that can process the neuroimages in the native space will be investigated in the future work. We also plan to examine the generalizability of the model on multi-center data acquired from different populations at different sites and scanners, as well as with different underlying diseases.   

%%%%%%%%%%%%%%%%%%%%%%%%%%%%%%%%%
\section{Conclusion}
\label{sec5}

PET imaging of CBF plays a key role in the assessment of cerebrovascular diseases. PET, however, is not readily accessible because of its prohibitive cost and use of ionizing radiation. This study introduces an attention network with a convolutional encoder-decoder structure to efficiently synthesize PET CBF maps from multi-contrast MRI scans without using radioactive tracers. The performance of the proposed image-to-image translation network is examined for different network settings and input MRI sequence combinations. Quantitative evaluations show improved PET synthesis results compared to previous MRI-to-PET CBF prediction models. Qualitative results also reveal that regional CBF values in synthetic PET are strongly correlated with those in ground-truth PET, with no statistically significant difference between them. In patients with cerebrovascular diseases, territories with abnormally low CBF were accurately identified in synthetic PET CBF maps. MRI-to-PET translation methods hold great potential for increasing the accessibility of cerebrovascular disease assessment for underserved populations, underprivileged communities, and developing nations. We hope this effort will drive further interest in the neuroimaging community to better leverage attention-based encoder-decoder networks and improve upon their current limitations, particularly reducing their computational cost.

%%%%%%%%%%%%%%%%%%%%%%%%%%%%%%%%%
\section*{Acknowledgments}
This work was supported by GE Healthcare, NIH, and the Stanford ADRC. Dr.~Ramy Hussein has received grant funding from NIH/NIA (P30 AG066515). Dr.~Moss Zhao is supported by the American Heart Association (Grant: 826254). Dr.~Greg Zaharchuk has received grant funding from NIH (R01-EB025220). 

\section*{Appendices}

\begin{figure*}[!ht]
	\centering
	\includegraphics[width=0.8\textwidth]{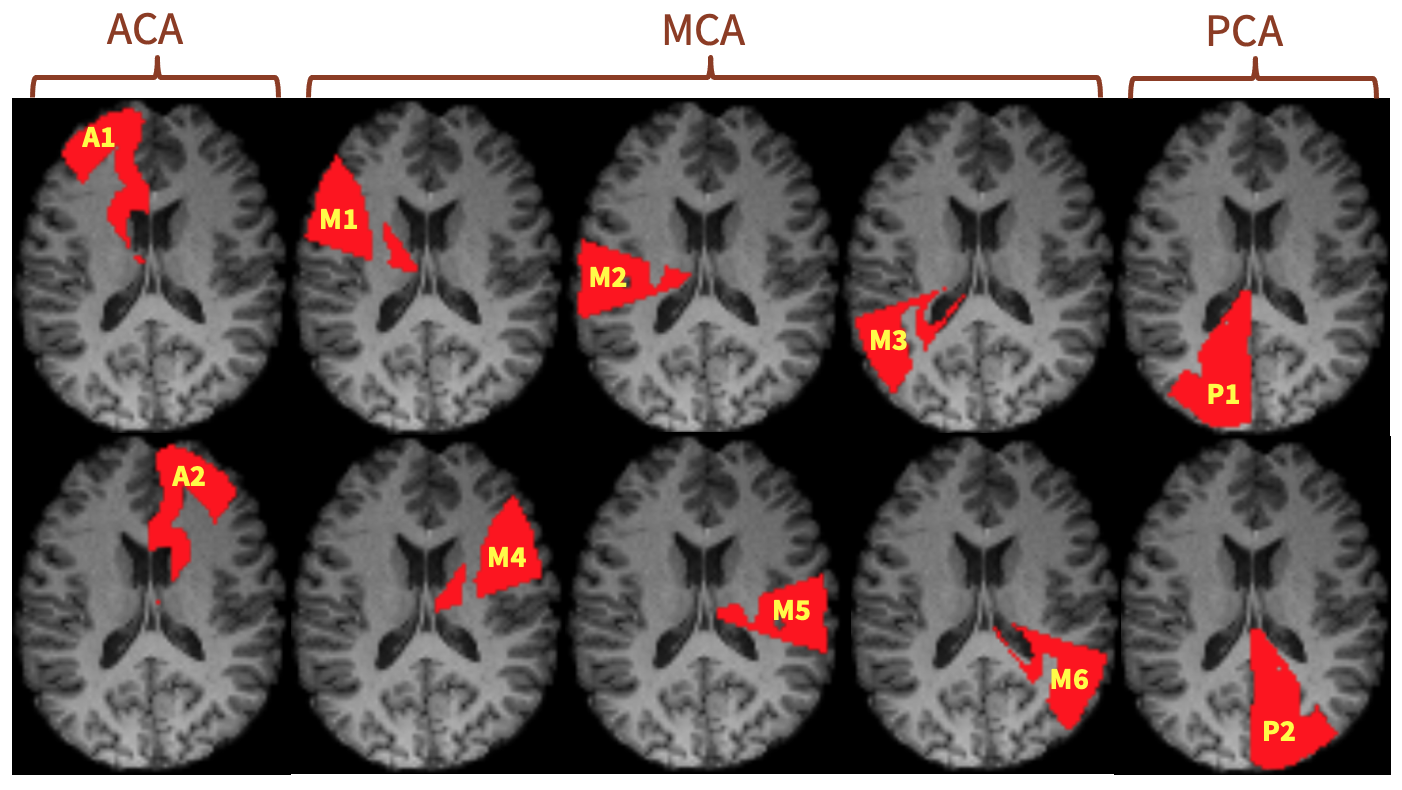}
	\caption{Brain arterial vascular territories of ASPECTS.}
	\label{fig.aspects}
\end{figure*}

Figure~\ref{fig.aspects} shows the 10 regions of interest (ROIs) of the ASPECTS mask. Each hemisphere has 5 ROIs: one anterior cerebral artery (ACA), three middle cerebral artery (MCA) and one posterior cerebral artery (PCA).

% \newpage

\medskip
{
	\small
	\bibliographystyle{IEEEtran}
	\bibliography{references}
}

\end{document}